%% file: aegis_paper_2026.tex
\documentclass[conference]{IEEEtran}

\usepackage{cite}
\usepackage{amsmath,amssymb,amsfonts}
\usepackage{booktabs}
\usepackage{graphicx}
\usepackage{adjustbox}
\usepackage{multirow}
\usepackage{textcomp}
\usepackage[hyphens]{url}
\usepackage{xcolor}
\usepackage{tikz}
\usepackage{pgf}
\usetikzlibrary{positioning}
\usetikzlibrary{shapes.geometric, arrows.meta, positioning, fit, calc}
\usepackage{placeins}       %

\def\BibTeX{{\rm B\kern-.05em{\sc i\kern-.025em b}\kern-.08em
    T\kern-.1667em\lower.7ex\hbox{E}\kern-.125emX}}

\newif\ifwithappendices%
\ifdefined\WITHAPPENDICES%
    \withappendicestrue%
\else
    \withappendicesfalse%
\fi

\newcommand{\titletext}{AEGIS: A Multi-Task Joint-Embedding Predictive Architecture for Mammography}
\title{\titletext}
\author{\IEEEauthorblockN{Scott Chase Waggener, Sai Karthik Navuluru, and Lakshman Tamil}
\IEEEauthorblockA{\textit{Department of Electrical and Computer Engineering} \\
\textit{University of Texas at Dallas}\\
Richardson, TX \\
scott.waggener@utdallas.edu; SaiKarthik.Navuluru@UTDallas.edu; laxman@utdallas.edu}
}

\begin{document}
\input{tikz/macros}
\maketitle

\begin{abstract}
We present Aegis, a joint-embedding predictive architecture for breast cancer detection
and density assessment in mammography. We train three Vision Transformer variants
(Small/Base/Large) using self-supervised joint-embedding predictive architecture
(JEPA) pre-training on 71,103 studies from 14 clinical sites, followed by
supervised fine-tuning with progressive resolution scaling up to
2048$\times$1536. On a curated 785-study test set, our largest model achieves
area under the receiver operating characteristic curve (AUC) 0.949 for breast
cancer triage with 93\% sensitivity and 75\% specificity at the optimal
operating point. An ensemble combining our model with a U.S. Food and Drug
Administration-cleared baseline further improves discrimination to 0.952 AUC.
For breast density classification, the model achieves 0.953 AUC for binary
(dense vs.\ non-dense) classification and 62.6\% exact accuracy across four
Breast Imaging Reporting and Data System (BI-RADS) categories, with 98.8\%
adjacent accuracy comparable to reported human inter-reader agreement.
External validation on the public VinDr-Mammo dataset provides evidence of cross-population transfer
under a different reference standard, with the largest model achieving 0.871 AUC
for triage in a zero-shot setting.
\end{abstract}


\section{Introduction}

Breast cancer remains one of the most prevalent and deadly malignancies worldwide, posing a significant public health challenge. In 2022, approximately 2.3 million women were diagnosed with breast cancer globally, resulting in an estimated 670,000 deaths \cite{bray2024global}.
This disease disproportionately affects women in lower-resource settings, where access to early detection and treatment is limited, leading to higher mortality rates; notably, the World Health Organization (WHO) reports that approximately 80\% of breast cancers occur in women with no specific risk factors other than sex and age \cite{WHO2025}, underscoring the limits of risk-factor-based prevention alone.
Incidence rates vary by region and socioeconomic development, with projections indicating a 38\% increase in cases and a 68\% rise in deaths by 2050 if current trends persist \cite{kim2025global}. 
The harm extends beyond mortality, encompassing treatment-related complications and long-term survivorship burdens for patients, families, and healthcare systems.

Artificial intelligence (AI) offers a promising path to address these challenges by augmenting radiologist capabilities in breast cancer screening.
Recent studies have demonstrated that AI systems can improve cancer detection rates while reducing false positives in clinical mammography workflows.
AI also enables automated breast density assessment and risk stratification, tasks traditionally subject to substantial inter-reader variability among radiologists.
Section~\ref{sec:related} reviews prior work on AI for mammography and the self-supervised learning methods that motivate our approach.

Joint-Embedding Predictive Architecture (JEPA) represents an advanced self-supervised learning paradigm designed to learn semantic representations from data without relying on hand-crafted augmentations or generative reconstruction.
In the image-based JEPA (I-JEPA) variant \cite{assran2023self}, the model predicts latent representations of target image blocks from a context encoder, promoting abstract, predictive understanding of visual structures while avoiding pitfalls like collapsed representations common in contrastive methods.
In this paper, we leverage JEPA-pretrained vision transformers to advance performance in breast cancer triage, breast density assessment, and lesion detection, demonstrating their efficacy on a large-scale mammographic dataset.

Our main contributions are:
\begin{itemize}
    \item A JEPA-based architecture adapted for mammography with CLS tokens for global classification and multi-task heads for triage, detection, and density assessment
    \item Demonstration that JEPA pre-training enables training Vision Transformers from scratch on a moderately-sized mammography dataset, achieving performance comparable to data-efficient convolutional neural network architectures without requiring ImageNet pre-training
    \item Comprehensive evaluation on a curated test set with biopsy-confirmed ground truth, including density assessment contextualized against human inter-reader agreement, and external validation on the public VinDr-Mammo benchmark under a different reference standard
\end{itemize}

\section{Related Work}
\label{sec:related}

\subsection{Self-Supervised Learning in Computer Vision}

Self-supervised learning has emerged as a powerful paradigm for learning visual representations without manual annotation.
Contrastive methods such as SimCLR~\cite{chen2020simple} and MoCo~\cite{he2019momentum} learn invariances by aligning augmented views while using negative examples or queues; non-contrastive bootstrap methods such as BYOL~\cite{grill2020bootstrap} instead use online and target networks to predict representations across augmented views without negative pairs.
While effective, these approaches depend heavily on carefully designed augmentation strategies, which may not transfer well to medical imaging domains where standard augmentations (color jittering, aggressive cropping) can destroy clinically relevant information.

Masked prediction approaches offer an alternative by reconstructing masked portions of the input.
Masked Autoencoders (MAE)~\cite{he2021masked} reconstruct pixel values of masked patches, BEiT~\cite{bao2021beit} predicts discrete visual tokens from a discrete-VAE tokenizer/codebook, and iBOT~\cite{zhou2021ibot} performs masked patch-token self-distillation with an online tokenizer.
These methods learn strong representations, but BEiT-style codebook supervision can require a pretrained in-domain tokenizer, while pixel or token reconstruction can emphasize low-level content rather than higher-level semantic features important for classification tasks.

Joint-embedding methods avoid explicit reconstruction by predicting in a learned latent space.
DINO~\cite{caron2021emerging} and DINOv2~\cite{oquab2023dinov} use self-distillation with a momentum teacher to learn semantic features that emerge without explicit supervision.
I-JEPA~\cite{assran2023self} extends this approach by predicting latent representations of target patches from context, avoiding both augmentation dependence and pixel-level reconstruction.
V-JEPA~\cite{bardes2024revisiting} demonstrates that this predictive architecture scales effectively to video understanding, while LeJEPA~\cite{balestriero2025lejepaprovablescalableselfsupervised} proposes a theoretically grounded JEPA objective with Sketched Isotropic Gaussian Regularization and removes several common implementation heuristics, including stop-gradient, teacher-student updates, and hyperparameter schedulers.
Our work builds on I-JEPA, adapting the joint-embedding predictive framework for mammography with domain-specific modifications including CLS tokens for global classification and multi-task supervised probes.

\subsection{Vision Transformers for Medical Imaging}

The Vision Transformer (ViT)~\cite{dosovitskiy2021imageworth16x16words} has demonstrated strong performance on natural image benchmarks but historically required large-scale pre-training datasets.
Data-efficient training strategies have addressed this limitation: DeiT~\cite{touvron2021trainingdataefficientimagetransformers} introduced knowledge distillation from convolutional neural network (CNN) teachers, while Compact Transformers~\cite{hassani2022escapingbigdataparadigm} proposed architectural modifications for smaller datasets.
For medical imaging specifically, self-supervised pre-training provides an effective path to leverage abundant unlabeled clinical data~\cite{shamshad2023transformers}.

Architectural improvements have enhanced ViT reliability.
Register tokens~\cite{darcet2024visiontransformersneedregisters} address attention artifacts that appear in deeper networks, improving feature quality for downstream tasks.
We incorporate these advances alongside modern transformer components: RoPE and SwiGLU from DINOv3~\cite{simeoni2025dinov3}, and RMSNorm~\cite{zhang2019rootmeansquarelayer} for normalization, adapting them for high-resolution mammographic images with progressive resolution scaling.

Despite these architectural advances, applying vision transformers to full-field digital mammography presents computational challenges.
Standard ViTs exhibit quadratic attention complexity with respect to sequence length; a $2048 \times 1536$ mammogram produces over 12,000 patches at $16 \times 16$ resolution, making direct processing prohibitive.
Consequently, most ViT-based mammography studies have relied on downsampled images~\cite{ayana2023vit}, hierarchical architectures with windowed attention such as the Swin Transformer~\cite{kassis2024dbt}, or hybrid CNN-ViT designs that extract features at multiple scales~\cite{kashiwada2025mimeViT}.

Our approach addresses these constraints through progressive resolution scaling during both pre-training ($256 \times 192 \rightarrow 512 \times 384$) and fine-tuning ($512 \times 384 \rightarrow 1024 \times 768 \rightarrow 2048 \times 1536$), enabling the model to learn representations at increasing fidelity while maintaining tractable training costs.
RoPE positional encoding further enables generalization to unseen resolutions without architectural modifications.

\subsection{AI Systems for Mammography}

Deep learning has shown substantial promise for breast cancer screening.
McKinney et al.~\cite{mckinney2020international} demonstrated that a CNN-based system could match or exceed radiologist performance on screening mammography, reducing false positives and false negatives across US and UK populations.
Subsequent real-world deployment studies have validated these findings at scale: Eisemann et al.~\cite{eisemann2025nationwide} reported that AI-assisted screening improved cancer detection rates by 17\% in a nationwide German implementation.

AI systems increasingly address multiple mammography tasks beyond cancer detection.
Automated breast density assessment provides consistent evaluations that support risk-stratified screening protocols~\cite{doi:10.2214/AJR.18.20391}, addressing the substantial inter-reader variability observed among radiologists~\cite{sprague2016variation}.
Risk prediction models integrate imaging features with clinical data to identify high-risk individuals years before diagnosis~\cite{doi:10.1148/ryai.230462}.
Public benchmarks such as VinDr-Mammo~\cite{nguyen2023vindr} enable standardized evaluation across research groups.

Our work differs from prior approaches by combining self-supervised JEPA pre-training with multi-task fine-tuning, simultaneously addressing triage, detection, and density assessment within a unified architecture.
This multi-task formulation enables shared representation learning across complementary clinical tasks while providing multiple outputs relevant to clinical workflow.

\section{Methods}

\paragraph{Architecture}
A standard Vision Transformer (ViT) \cite{dosovitskiy2021imageworth16x16words} is adopted as the backbone model and modernized to more closely
resemble the design used by DINOv3 \cite{simeoni2025dinov3}. Notably, we adopt rotary position encoding (RoPE) \cite{su2023roformer} as used by DINOv3,
incorporate SwiGLU \cite{shazeer2020gluvariantsimprovetransformer} as the activation function and RMSNorm \cite{zhang2019rootmeansquarelayer} as the normalization layer,
and include learnable register tokens \cite{darcet2024visiontransformersneedregisters}.

We model our JEPA after I-JEPA \cite{assran2023self} and incorporate several modifications used by CAPI \cite{chen2024capi} and DINOv3 \cite{simeoni2025dinov3}.
Our JEPA predictor is a cross-attention module (without self-attention) that shares hyperparameters (hidden size, activation, etc.) with the backbone.
Target queries are initialized using learnable positional embeddings.
Context and target masks are generated using a constant fraction of non-overlapping tokens in contiguous blocks with a parameterized scale.
When scaling resolution later in training the scale of the contiguous blocks is increased according to the resolution scaling factor.
Gram anchoring is applied to the student's output embeddings to improve dense feature quality.

In a departure from I-JEPA, we introduce \texttt{CLS} tokens to the backbone with the aim of learning strong global representations.
To train these tokens a second forward pass of the predictor is performed using the \texttt{CLS} tokens as context in place of the visual tokens.
A SigREG loss \cite{balestriero2025lejepaprovablescalableselfsupervised} is applied to the student's output \texttt{CLS} tokens to increase
their suitability for downstream probing tasks.

To fine-tune the model for downstream tasks we adopt distinct heads for triage, detection, and density estimation tasks.
Triage and density estimation use binary classification heads, with density labels being mapped from categorical values
to the range $[0, 1]$ according to:
\begin{equation}
\text{Density} = \begin{cases}
0.0 & \text{if } \text{BI-RADS} = A \\
0.33 & \text{if } \text{BI-RADS} = B \\
0.66 & \text{if } \text{BI-RADS} = C \\
1.0 & \text{if } \text{BI-RADS} = D
\end{cases}
\end{equation}
Detection is achieved in a similar fashion to CenterNet \cite{zhou2019objects}, wherein bounding boxes are mapped to Gaussian splats centered within the bounding box.
Since Aegis is not intended to output bounding boxes, we forego the use of a separate detection head regressing the bounding box coordinates and focus exclusively on heatmap prediction.

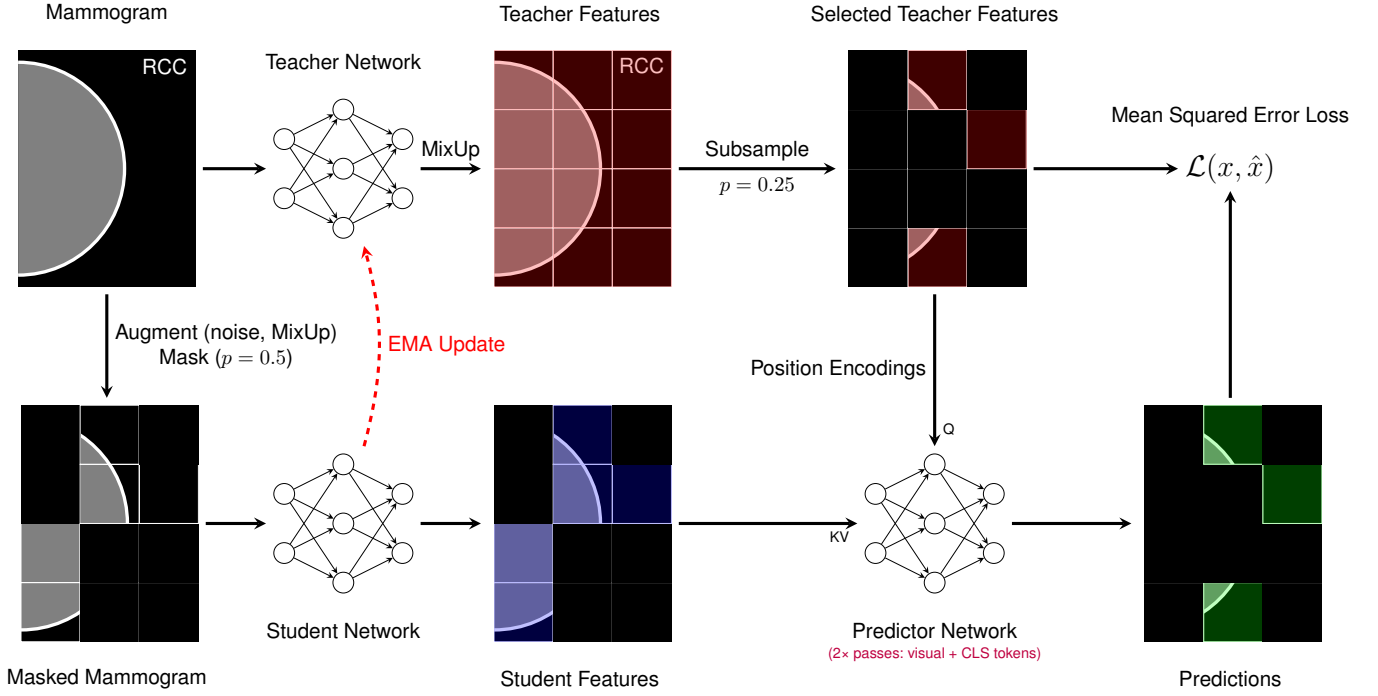
\begin{figure*}[t]
\centering
\input{tikz/jepa.tex}
\caption{
    JEPA pre-training architecture.
    A student network processes masked mammograms while a teacher network (updated via EMA) processes the unmasked input.
    The predictor network uses cross-attention to predict target embeddings from context embeddings, with position encodings as queries.
}
\label{fig:jepa}
\end{figure*}

\begin{figure*}[t]
\centering
\input{tikz/finetune.tex}
\caption{
    Supervised fine-tuning architecture. 
    The pre-trained ViT backbone processes mammograms and outputs a grid of visual tokens along with 4 CLS tokens. 
    Visual tokens are used for dense prediction tasks: cancerous lesion detection and tissue segmentation (3-class categorical: background, breast tissue, pectoral muscle).
    CLS tokens are used for a specific classification task:
    CLS0 for triage (binary), CLS1 for multi-class binary view classification (MLO, CC, Spot/Mag), CLS2 for breast implant detection (binary),
    and CLS3 for breast density estimation (continuous value in [0,1]).
}
\label{fig:finetune}
\end{figure*}
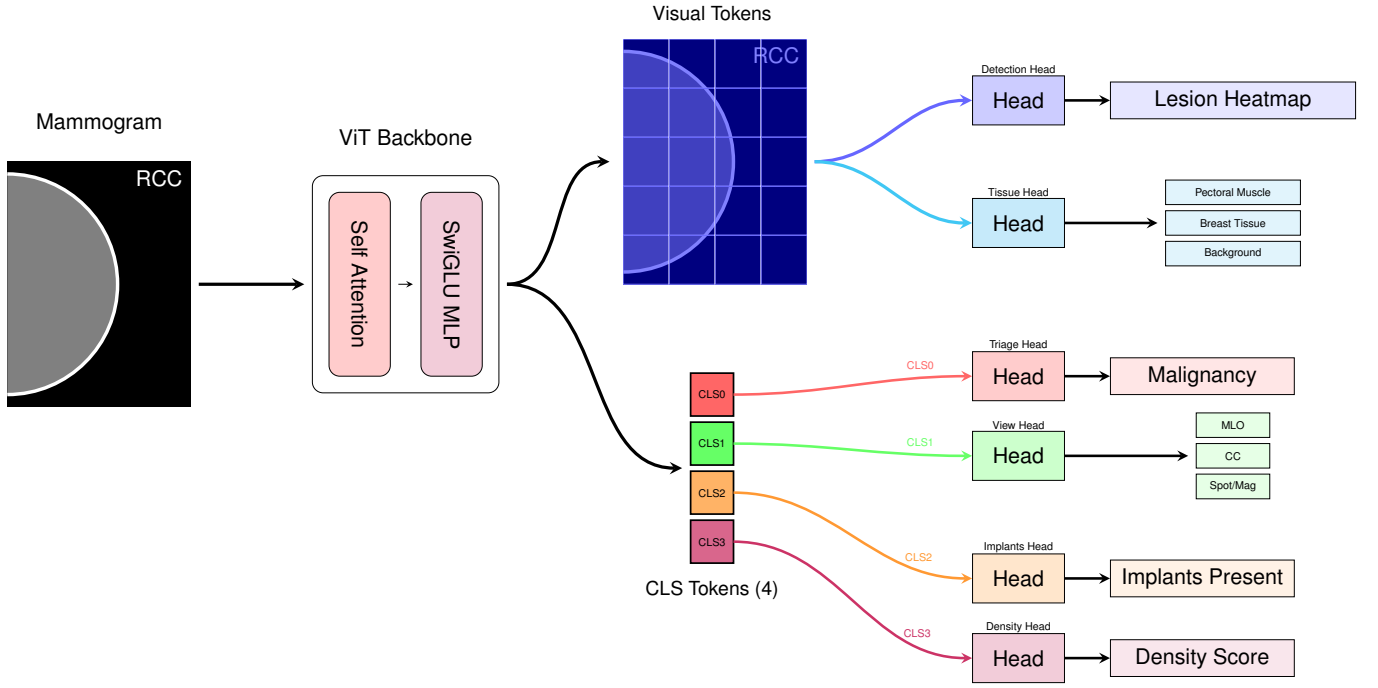

\paragraph{Training}
We train three model variants: ViT-Small/16, ViT-Base/16, and ViT-Large/16, following a two-stage training protocol.
Pre-training consists of 375 epochs combining the JEPA self-supervised objective with supervised auxiliary probes for triage, detection, density, view classification, and implant detection.
The view classification, implant detection, and tissue segmentation heads serve as auxiliary training objectives that improve representation learning; these outputs are not evaluated as clinical endpoints since view and implant information is typically available in DICOM metadata, and tissue segmentation has limited standalone clinical utility.
Training uses progressive resolution scheduling, starting at 256$\times$192 and increasing to 512$\times$384 for the final 75 epochs.
Fine-tuning continues for 45 epochs, adding tissue segmentation and risk prediction heads while progressively increasing resolution through three stages: 512$\times$384, 1024$\times$768, and finally 2048$\times$1536.
As resolution increases, batch size must decrease to avoid out-of-memory errors; we apply a proportionate increase in gradient accumulation steps to maintain a constant effective batch size.
All training was performed on 2$\times$ NVIDIA RTX5090 GPUs using distributed data parallel.

\input{tables/data_variables.tex}

\paragraph{Data}

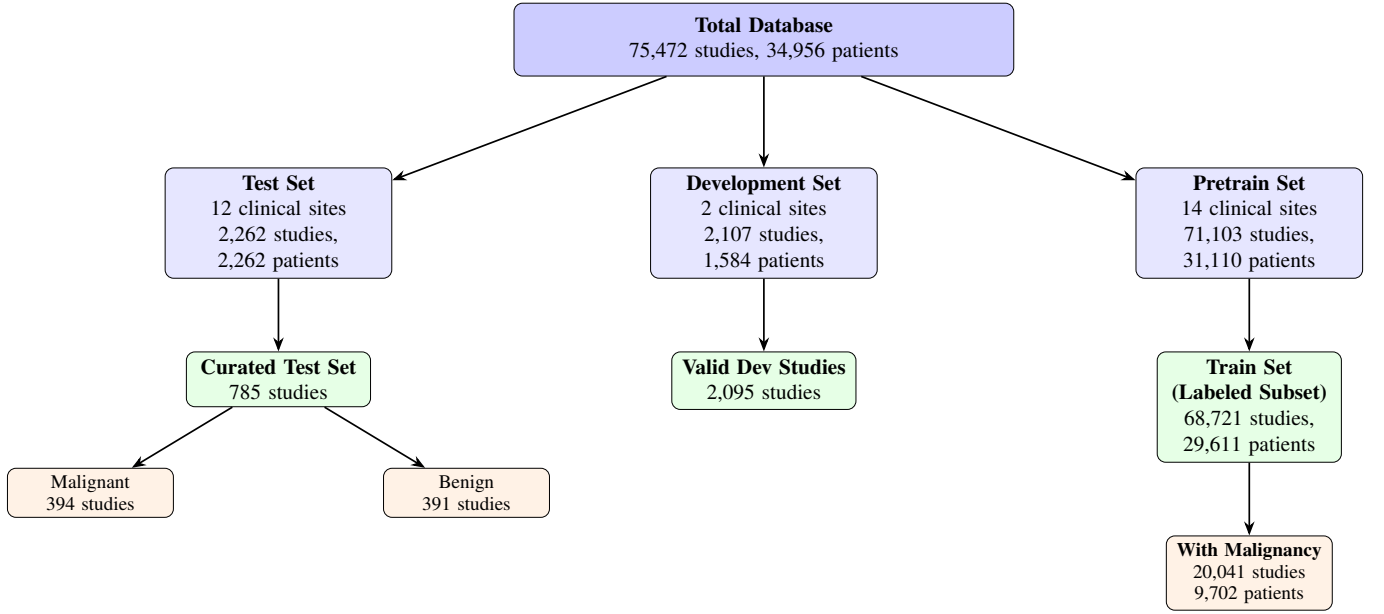
\begin{figure*}[t]
\centering
\resizebox{\textwidth}{!}{
\begin{tikzpicture}[
    node distance=1cm and 0.5cm,
    block/.style={rectangle, draw, fill=blue!10, text width=3.5cm, text centered, rounded corners, minimum height=1cm},
    subblock/.style={rectangle, draw, fill=green!10, text width=2.8cm, text centered, rounded corners, minimum height=0.8cm},
    smallblock/.style={rectangle, draw, fill=orange!10, text width=2.5cm, text centered, rounded corners, minimum height=0.7cm, font=\small},
    arrow/.style={-Stealth, thick}
]

\node[block, fill=blue!20, text width=8cm, minimum height=1.2cm] (total) {
    \textbf{Total Database}\\
    \totalStudies{} studies, \totalPatients{} patients
};

\node[block, below left=1.5cm and 2cm of total] (test) {
    \textbf{Test Set}\\
    \clinicalSitesTest{} clinical sites\\
    \totalStudiesTest{} studies, \totalPatientsTest{} patients
};

\node[block, below=1.5cm of total] (dev) {
    \textbf{Development Set}\\
    \clinicalSitesDev{} clinical sites\\
    \totalStudiesDev{} studies, \totalPatientsDev{} patients
};

\node[block, below right=1.5cm and 2cm of total] (pretrain) {
    \textbf{Pretrain Set}\\
    \clinicalSitesPretrain{} clinical sites\\
    \totalStudiesPretrain{} studies, \totalPatientsPretrain{} patients
};

\node[subblock, below=1.2cm of test] (curated) {
    \textbf{Curated Test Set}\\
    \curatedStudiesTest{} studies
};

\node[smallblock, below left=1cm and 0.2cm of curated] (malignant) {
    Malignant\\
    \curatedStudiesTestMalignant{} studies
};

\node[smallblock, below right=1cm and 0.2cm of curated] (benign) {
    Benign\\
    \curatedStudiesTestBenign{} studies
};

\node[subblock, below=1.2cm of dev] (validdev) {
    \textbf{Valid Dev Studies}\\
    \validStudiesDev{} studies
};

\node[subblock, below=1.2cm of pretrain] (train) {
    \textbf{Train Set (Labeled Subset)}\\
    \totalStudiesTrain{} studies, \totalPatientsTrain{} patients
};

\node[smallblock, below=1.2cm of train] (trainmain) {
    \textbf{With Malignancy}\\
    \totalStudiesTrainWithMalignancy{} studies\\
    \totalPatientsTrainWithMalignancy{} patients
};

\draw[arrow] (total) -- (test);
\draw[arrow] (total) -- (dev);
\draw[arrow] (total) -- (pretrain);
\draw[arrow] (test) -- (curated);
\draw[arrow] (curated) -- (malignant);
\draw[arrow] (curated) -- (benign);
\draw[arrow] (dev) -- (validdev);
\draw[arrow] (pretrain) -- (train);
\draw[arrow] (train) -- (trainmain);

\end{tikzpicture}
}
\caption{Overview of data partitioning in the MedCognetics mammography database. The pretrain set is used for self-supervised learning and contains the train set as a labeled subset for supervised fine-tuning. Test and development sets are held out from separate clinical sites. The curated test set is sampled to represent a real-world screening population. Valid dev studies exclude incomplete studies (e.g., digital breast tomosynthesis-only modality).}
\label{fig:data_partitions}
\end{figure*}

Aegis is trained and evaluated on partitions of the MedCognetics proprietary mammography database.
The database is partitioned into pretrain, train, development, and test sets.
The pretrain set (\totalStudiesPretrain\ studies, \totalPatientsPretrain\ patients, \clinicalSitesPretrain\ clinical sites) 
is used for self-supervised JEPA pretraining and is a superset containing the train partition.
The train partition (\totalStudiesTrain\ studies, \totalPatientsTrain\ patients) is a labeled subset of the pretrain set 
used for supervised fine-tuning, with \totalStudiesTrainWithMalignancy\ studies containing malignancy annotations.
We withhold \clinicalSitesTest\ clinical sites for testing and \clinicalSitesDev\ clinical sites for validation.
An overview of the data partitions is shown in Figure \ref{fig:data_partitions} and Table \ref{tab:data_partition_summary}.

Reference standard for test data was established by biopsy confirmation for malignant studies or two year stability for benign studies.
Breast density ground truth was provided by a single Mammography Quality Standards
Act (MQSA)-certified mammographer.

\paragraph{External Validation Dataset}
We evaluate external generalization on VinDr-Mammo~\cite{nguyen2023vindr}, a publicly available dataset of 5,000 full-field digital mammography (FFDM) exams acquired from a hospital in Vietnam.
VinDr-Mammo exams were independently double-read by experienced radiologists, with discordant assessments resolved by arbitration from a third radiologist.
We use the 1,000-exam test split, classifying studies with adjudicated BI-RADS 4--5 assessments as positive and BI-RADS 1--2 assessments as negative; BI-RADS 3 studies are excluded due to their indeterminate nature, yielding 909 evaluable studies (96 positive, 813 negative).
Models are applied in a zero-shot transfer setting: no fine-tuning is performed on VinDr-Mammo data, and classification thresholds determined on the proprietary development set are applied unchanged.
Unlike the proprietary test set where malignancy is confirmed by biopsy or two-year stability, VinDr-Mammo ground truth reflects adjudicated radiologist BI-RADS assessments without pathological confirmation, making direct comparison of absolute AUC values inappropriate.

\input{tables/data_partition_summary.tex}

\paragraph{Baseline Model}
We compare Aegis against a U.S. Food and Drug Administration-cleared
convolutional neural network (CNN) for breast cancer triage.
This baseline model outputs only study-level malignancy scores and lacks a detection head.
The CNN is excluded from density evaluation as it outputs constant density values.

\paragraph{Evaluation}
Quantitative evaluation focuses on breast cancer triage and breast density assessment.
Optimal classification thresholds are determined using Youden's J statistic~\cite{youden1950index} ($J = \text{Sensitivity} + \text{Specificity} - 1$) on the development set and applied unchanged to the test set.
For triage evaluation, we compute three score types:
(1) \emph{Triage} score from the classification head alone,
(2) \emph{Triage+Detection} score averaging triage with max-pooled detection output, and
(3) \emph{Ensemble} score combining ViT Triage+Detection with CNN Triage via simple averaging.
The Triage+Detection combination improves discrimination by incorporating lesion-level detection features into the study-level classification.
Detection performance is evaluated using study-level AUC from the max-pooled heatmap score rather than localization metrics such as mean average precision (mAP) or intersection-over-union (IoU); this approach reflects clinical practice where the presence of a suspicious finding matters more than precise boundary delineation, and accounts for inherent ambiguity in lesion boundaries where radiologists often disagree on exact extent.
All metrics are reported with 95\% bootstrap confidence intervals (1000 samples).
The curated test set size (N=785) and 50\% prevalence were determined via power analysis~\cite{obuchowski2004roc} targeting AUC of 0.95 at significance level 0.05, enriched relative to an annual U.S. female breast cancer incidence rate on the order of 127 cases per 100,000 women (approximately 0.127\% per year)~\cite{cdc2023cancer}.
The curated test set consists of studies acquired via digital breast tomosynthesis (DBT); performance is evaluated on the corresponding 2D modality (FFDM or synthesized 2D views).

\input{results/combined/variables.tex}

\input{results/combined/delong/variables.tex}

\input{results/vit-l/test/density/multiclass/latex/variables.tex}
\input{results/vit-l/test/density/binary/latex/variables.tex}
\input{results/vit-b/test/density/multiclass/latex/variables.tex}
\input{results/vit-b/test/density/binary/latex/variables.tex}
\input{results/vit-s/test/density/multiclass/latex/variables.tex}
\input{results/vit-s/test/density/binary/latex/variables.tex}

\input{results/vit-l/vindr/density/multiclass/latex/variables.tex}
\input{results/vit-l/vindr/density/binary/latex/variables.tex}
\input{results/vit-b/vindr/density/multiclass/latex/variables.tex}
\input{results/vit-b/vindr/density/binary/latex/variables.tex}
\input{results/vit-s/vindr/density/multiclass/latex/variables.tex}
\input{results/vit-s/vindr/density/binary/latex/variables.tex}

\section{Results}

We evaluate Aegis across three ViT model scales (Small, Base, Large) and compare against a CNN baseline on breast cancer triage and density classification tasks.
All performance metrics are computed using thresholds optimized on the development set via Youden's J statistic and applied unchanged to the test set.

\subsection{Triage Performance}

Table~\ref{tab:dev_auc_comparison} compares discrimination performance across score types on the development set (N=\vitlDevTriageN{} studies).
The detection head alone achieves strong discrimination (ViT-L: \vitlDevDetectionAuc{}), demonstrating that the max-pooled heatmap score captures meaningful lesion-level information.
Combining triage and detection scores (Triage+Detection) improves AUC for all ViT variants, with ViT-L achieving \vitlDevTriagedetectionAuc{} (95\% CI: \vitlDevTriagedetectionAucLb{}--\vitlDevTriagedetectionAucUb{}).
The ensemble combining ViT Triage+Detection with CNN Triage further improves discrimination, achieving AUC of \vitlDevEnsembleAuc{} for ViT-L.
Development set thresholds determined via Youden's J statistic are applied unchanged to the test set.

\begin{table*}[t]
\centering
\caption{AUC comparison across score types on the development set. Triage uses the classification head alone; Detection uses the max-pooled heatmap score; Triage+Detection averages both; Ensemble combines ViT Triage+Detection with CNN Triage. N=\vitlDevTriageN{} studies (\vitlDevTriageNPos{} malignant, \vitlDevTriageNNeg{} benign).}
\label{tab:dev_auc_comparison}
\begin{adjustbox}{max width=\textwidth}
\begin{tabular}{lrrrr}
\toprule
\textbf{Model} & \textbf{Triage} & \textbf{Detection} & \textbf{Triage+Det.} & \textbf{Ensemble} \\
\midrule
ViT-L & \vitlDevTriageAuc{} [\vitlDevTriageAucLb{}--\vitlDevTriageAucUb{}] & \vitlDevDetectionAuc{} [\vitlDevDetectionAucLb{}--\vitlDevDetectionAucUb{}] & \vitlDevTriagedetectionAuc{} [\vitlDevTriagedetectionAucLb{}--\vitlDevTriagedetectionAucUb{}] & \vitlDevEnsembleAuc{} [\vitlDevEnsembleAucLb{}--\vitlDevEnsembleAucUb{}] \\
ViT-B & \vitbDevTriageAuc{} [\vitbDevTriageAucLb{}--\vitbDevTriageAucUb{}] & \vitbDevDetectionAuc{} [\vitbDevDetectionAucLb{}--\vitbDevDetectionAucUb{}] & \vitbDevTriagedetectionAuc{} [\vitbDevTriagedetectionAucLb{}--\vitbDevTriagedetectionAucUb{}] & \vitbDevEnsembleAuc{} [\vitbDevEnsembleAucLb{}--\vitbDevEnsembleAucUb{}] \\
ViT-S & \vitsDevTriageAuc{} [\vitsDevTriageAucLb{}--\vitsDevTriageAucUb{}] & \vitsDevDetectionAuc{} [\vitsDevDetectionAucLb{}--\vitsDevDetectionAucUb{}] & \vitsDevTriagedetectionAuc{} [\vitsDevTriagedetectionAucLb{}--\vitsDevTriagedetectionAucUb{}] & \vitsDevEnsembleAuc{} [\vitsDevEnsembleAucLb{}--\vitsDevEnsembleAucUb{}] \\
CNN Baseline & \CnnDevTriageAuc{} [\CnnDevTriageAucLb{}--\CnnDevTriageAucUb{}] & --- & --- & --- \\
\bottomrule
\end{tabular}
\end{adjustbox}
\end{table*}

Table~\ref{tab:dev_triage_comparison} reports sensitivity and specificity at the optimal operating point for the Triage+Detection score (CNN uses Triage only).

\begin{table*}[t]
\centering
\caption{Sensitivity and specificity at optimal operating points on the development set using Triage+Detection score (CNN: Triage only). Thresholds determined via Youden's J statistic. N=\vitlDevTriageN{} studies (\vitlDevTriageNPos{} malignant, \vitlDevTriageNNeg{} benign).}
\label{tab:dev_triage_comparison}
\begin{adjustbox}{max width=\textwidth}
\begin{tabular}{lrr}
\toprule
\textbf{Model} & \textbf{Sensitivity [95\% CI]} & \textbf{Specificity [95\% CI]} \\
\midrule
ViT-L & \vitlDevTriagedetectionSens{}\% [\vitlDevTriagedetectionSensLb{}--\vitlDevTriagedetectionSensUb{}] & \vitlDevTriagedetectionSpec{}\% [\vitlDevTriagedetectionSpecLb{}--\vitlDevTriagedetectionSpecUb{}] \\
ViT-B & \vitbDevTriagedetectionSens{}\% [\vitbDevTriagedetectionSensLb{}--\vitbDevTriagedetectionSensUb{}] & \vitbDevTriagedetectionSpec{}\% [\vitbDevTriagedetectionSpecLb{}--\vitbDevTriagedetectionSpecUb{}] \\
ViT-S & \vitsDevTriagedetectionSens{}\% [\vitsDevTriagedetectionSensLb{}--\vitsDevTriagedetectionSensUb{}] & \vitsDevTriagedetectionSpec{}\% [\vitsDevTriagedetectionSpecLb{}--\vitsDevTriagedetectionSpecUb{}] \\
CNN Baseline & \CnnDevTriageSens{}\% [\CnnDevTriageSensLb{}--\CnnDevTriageSensUb{}] & \CnnDevTriageSpec{}\% [\CnnDevTriageSpecLb{}--\CnnDevTriageSpecUb{}] \\
\bottomrule
\end{tabular}
\end{adjustbox}
\end{table*}

\begin{figure*}[t]
\centering
\includegraphics[width=0.7\textwidth]{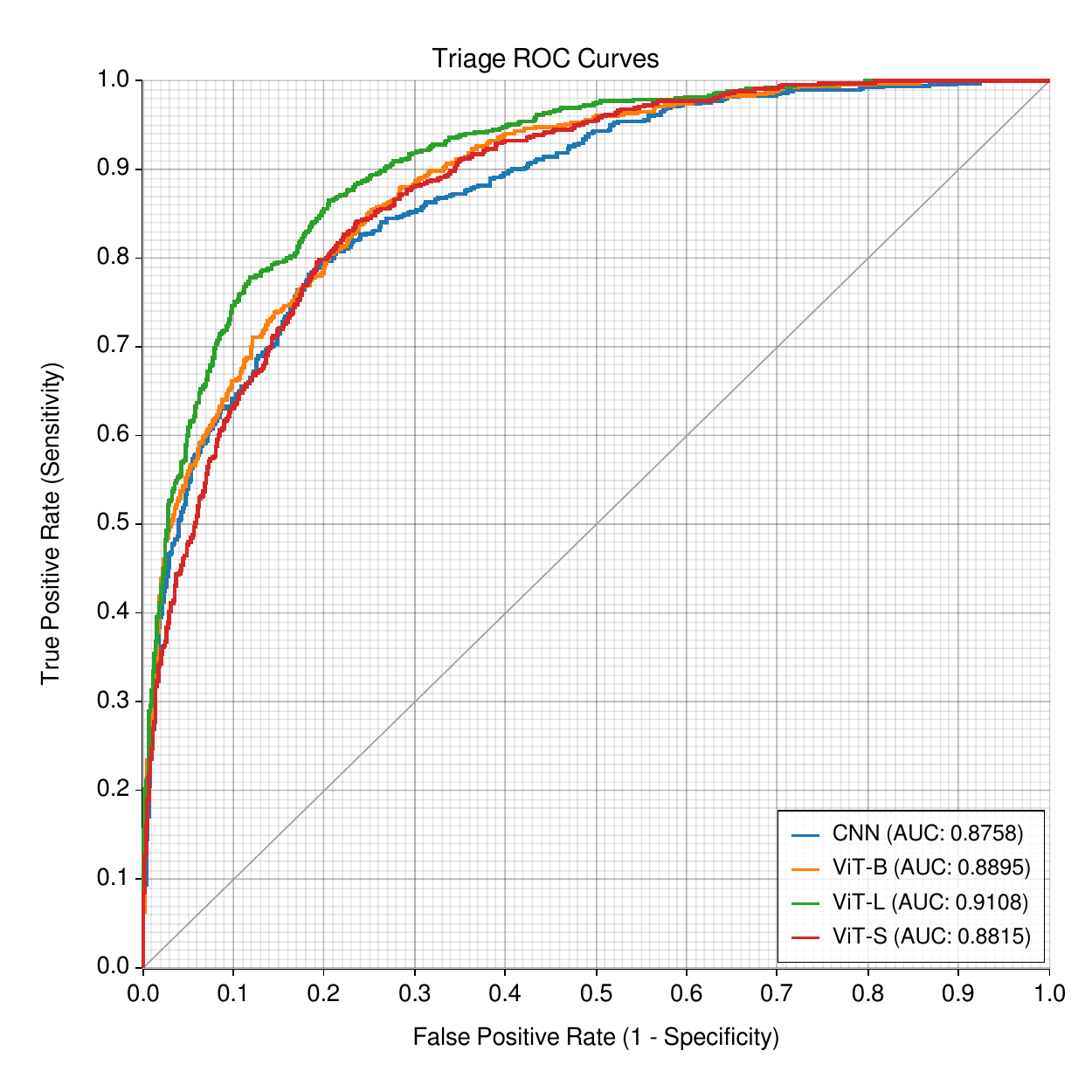}
\caption[Breast cancer triage curves on the development set]{Receiver operating characteristic (ROC) curves for breast cancer triage (Triage+Detection score; CNN: Triage only) on the development set. Operating points indicate the threshold selected via Youden's J statistic.}
\label{fig:roc_dev}
\end{figure*}

Table~\ref{tab:fda_auc_comparison} presents AUC comparison across score types on the test set (N=\vitlTestfdaTriageN{} DBT studies, evaluated on the corresponding 2D modality).
The detection score alone achieves AUC of \vitlTestfdaDetectionAuc{} for ViT-L, comparable to the triage score (\vitlTestfdaTriageAuc{}).
Combining both scores (Triage+Detection) yields the best single-model performance with AUC of \vitlTestfdaTriagedetectionAuc{} (95\% CI: \vitlTestfdaTriagedetectionAucLb{}--\vitlTestfdaTriagedetectionAucUb{}).
The ensemble further improves to \vitlTestfdaEnsembleAuc{}.

\begin{table*}[t]
\centering
\caption{AUC comparison across score types on the test set. Triage uses the classification head alone; Detection uses the max-pooled heatmap score; Triage+Detection averages both; Ensemble combines ViT Triage+Detection with CNN Triage. N=\vitlTestfdaTriageN{} studies (\vitlTestfdaTriageNPos{} malignant, \vitlTestfdaTriageNNeg{} benign).}
\label{tab:fda_auc_comparison}
\begin{adjustbox}{max width=\textwidth}
\begin{tabular}{lrrrr}
\toprule
\textbf{Model} & \textbf{Triage} & \textbf{Detection} & \textbf{Triage+Det.} & \textbf{Ensemble} \\
\midrule
ViT-L & \vitlTestfdaTriageAuc{} [\vitlTestfdaTriageAucLb{}--\vitlTestfdaTriageAucUb{}] & \vitlTestfdaDetectionAuc{} [\vitlTestfdaDetectionAucLb{}--\vitlTestfdaDetectionAucUb{}] & \vitlTestfdaTriagedetectionAuc{} [\vitlTestfdaTriagedetectionAucLb{}--\vitlTestfdaTriagedetectionAucUb{}] & \vitlTestfdaEnsembleAuc{} [\vitlTestfdaEnsembleAucLb{}--\vitlTestfdaEnsembleAucUb{}] \\
ViT-B & \vitbTestfdaTriageAuc{} [\vitbTestfdaTriageAucLb{}--\vitbTestfdaTriageAucUb{}] & \vitbTestfdaDetectionAuc{} [\vitbTestfdaDetectionAucLb{}--\vitbTestfdaDetectionAucUb{}] & \vitbTestfdaTriagedetectionAuc{} [\vitbTestfdaTriagedetectionAucLb{}--\vitbTestfdaTriagedetectionAucUb{}] & \vitbTestfdaEnsembleAuc{} [\vitbTestfdaEnsembleAucLb{}--\vitbTestfdaEnsembleAucUb{}] \\
ViT-S & \vitsTestfdaTriageAuc{} [\vitsTestfdaTriageAucLb{}--\vitsTestfdaTriageAucUb{}] & \vitsTestfdaDetectionAuc{} [\vitsTestfdaDetectionAucLb{}--\vitsTestfdaDetectionAucUb{}] & \vitsTestfdaTriagedetectionAuc{} [\vitsTestfdaTriagedetectionAucLb{}--\vitsTestfdaTriagedetectionAucUb{}] & \vitsTestfdaEnsembleAuc{} [\vitsTestfdaEnsembleAucLb{}--\vitsTestfdaEnsembleAucUb{}] \\
CNN Baseline & \CnnTestfdaTriageAuc{} [\CnnTestfdaTriageAucLb{}--\CnnTestfdaTriageAucUb{}] & --- & --- & --- \\
\bottomrule
\end{tabular}
\end{adjustbox}
\end{table*}

Table~\ref{tab:fda_triage_comparison} reports sensitivity and specificity at the optimal operating point for the Triage+Detection score.
ViT-L achieves \vitlTestfdaTriagedetectionSens{}\% sensitivity and \vitlTestfdaTriagedetectionSpec{}\% specificity, outperforming the CNN baseline.
DeLong's test~\cite{delong1988comparing} confirms that ViT-L significantly outperforms the CNN baseline on the Triage score
(AUC difference: \delongVitlVsCnnAucDiff{}, 95\% CI: \delongVitlVsCnnAucDiffLb{}--\delongVitlVsCnnAucDiffUb{}, p=\delongVitlVsCnnPval{}).
ViT-S and ViT-B do not show statistically significant differences from CNN (p=\delongVitsVsCnnPval{} and p=\delongVitbVsCnnPval{}, respectively).

\begin{table*}[t]
\centering
\caption{Sensitivity and specificity at optimal operating points on the test set using Triage+Detection score (CNN: Triage only). Thresholds determined via Youden's J on the development set. N=\vitlTestfdaTriageN{} studies (\vitlTestfdaTriageNPos{} malignant, \vitlTestfdaTriageNNeg{} benign).}
\label{tab:fda_triage_comparison}
\begin{adjustbox}{max width=\textwidth}
\begin{tabular}{lrr}
\toprule
\textbf{Model} & \textbf{Sensitivity [95\% CI]} & \textbf{Specificity [95\% CI]} \\
\midrule
ViT-L & \vitlTestfdaTriagedetectionSens{}\% [\vitlTestfdaTriagedetectionSensLb{}--\vitlTestfdaTriagedetectionSensUb{}] & \vitlTestfdaTriagedetectionSpec{}\% [\vitlTestfdaTriagedetectionSpecLb{}--\vitlTestfdaTriagedetectionSpecUb{}] \\
ViT-B & \vitbTestfdaTriagedetectionSens{}\% [\vitbTestfdaTriagedetectionSensLb{}--\vitbTestfdaTriagedetectionSensUb{}] & \vitbTestfdaTriagedetectionSpec{}\% [\vitbTestfdaTriagedetectionSpecLb{}--\vitbTestfdaTriagedetectionSpecUb{}] \\
ViT-S & \vitsTestfdaTriagedetectionSens{}\% [\vitsTestfdaTriagedetectionSensLb{}--\vitsTestfdaTriagedetectionSensUb{}] & \vitsTestfdaTriagedetectionSpec{}\% [\vitsTestfdaTriagedetectionSpecLb{}--\vitsTestfdaTriagedetectionSpecUb{}] \\
CNN Baseline & \CnnTestfdaTriageSens{}\% [\CnnTestfdaTriageSensLb{}--\CnnTestfdaTriageSensUb{}] & \CnnTestfdaTriageSpec{}\% [\CnnTestfdaTriageSpecLb{}--\CnnTestfdaTriageSpecUb{}] \\
\bottomrule
\end{tabular}
\end{adjustbox}
\end{table*}

\begin{figure*}[t]
\centering
\includegraphics[width=0.7\textwidth]{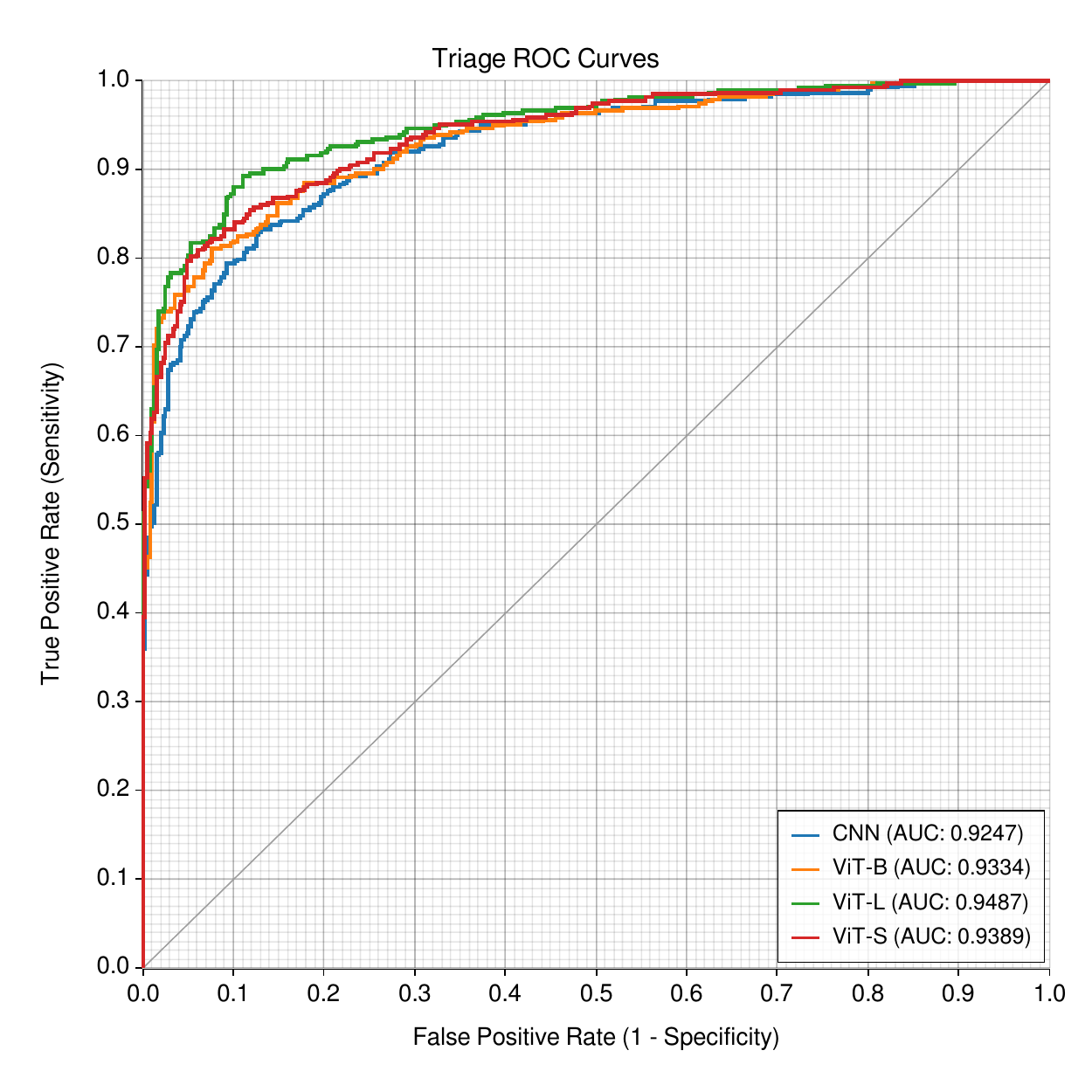}
\caption{ROC curves for breast cancer triage (Triage+Detection score; CNN: Triage only) on the test set. Operating points indicate the threshold selected via Youden's J statistic on the development set.}
\label{fig:roc_fda}
\end{figure*}

Figure~\ref{fig:roc_fda} shows the ROC curves for all models on the test set.

\subsection{Breast Density Classification}

Aegis provides automated breast density assessment as a secondary output.
Table~\ref{tab:density_comparison} compares density classification performance across all ViT model variants.
For binary classification (dense vs.\ non-dense), all models achieve AUC above 0.95, with ViT-L achieving \vitlDensityBinaryAuc{}
(95\% CI: \vitlDensityBinaryAucLb{}--\vitlDensityBinaryAucUb{}).

For multiclass classification across all four BI-RADS density categories (A/B/C/D),
ViT-L achieves an exact accuracy of \vitlDensityAccuracyMean{} (95\% CI: \vitlDensityAccuracyCILower{}--\vitlDensityAccuracyCIUpper{})
and adjacent accuracy of \vitlDensityAdjacentAccuracyMean{} (95\% CI: \vitlDensityAdjacentAccuracyCILower{}--\vitlDensityAdjacentAccuracyCIUpper{}).
The high adjacent accuracy indicates that when the model misclassifies density, predictions are typically within one category of the ground truth.

\begin{table*}[t]
\centering
\caption{Breast density classification performance comparison across ViT model variants on the test set, with published human clinical-assessment agreement shown for context. Binary classification distinguishes dense (C/D) from non-dense (A/B) tissue. Multiclass accuracy is exact match across all four BI-RADS categories; adjacent accuracy allows off-by-one errors.}
\label{tab:density_comparison}
\begin{adjustbox}{max width=\textwidth}
\begin{tabular}{lrrr}
\toprule
\textbf{Model} & \textbf{Binary AUC [95\% CI]} & \textbf{4-Class Acc./Agree.} & \textbf{Adjacent Acc./Agree.} \\
\midrule
ViT-L & \vitlDensityBinaryAuc{} [\vitlDensityBinaryAucLb{}--\vitlDensityBinaryAucUb{}] & \vitlDensityAccuracyMean{} [\vitlDensityAccuracyCILower{}--\vitlDensityAccuracyCIUpper{}] & \vitlDensityAdjacentAccuracyMean{} [\vitlDensityAdjacentAccuracyCILower{}--\vitlDensityAdjacentAccuracyCIUpper{}] \\
ViT-B & \vitbDensityBinaryAuc{} [\vitbDensityBinaryAucLb{}--\vitbDensityBinaryAucUb{}] & \vitbDensityAccuracyMean{} [\vitbDensityAccuracyCILower{}--\vitbDensityAccuracyCIUpper{}] & \vitbDensityAdjacentAccuracyMean{} [\vitbDensityAdjacentAccuracyCILower{}--\vitbDensityAdjacentAccuracyCIUpper{}] \\
ViT-S & \vitsDensityBinaryAuc{} [\vitsDensityBinaryAucLb{}--\vitsDensityBinaryAucUb{}] & \vitsDensityAccuracyMean{} [\vitsDensityAccuracyCILower{}--\vitsDensityAccuracyCIUpper{}] & \vitsDensityAdjacentAccuracyMean{} [\vitsDensityAdjacentAccuracyCILower{}--\vitsDensityAdjacentAccuracyCIUpper{}] \\
\midrule
Human$^\dagger$ & --- & 0.67--0.80 & $>$0.99 \\
\bottomrule
\end{tabular}
\end{adjustbox}
\\[0.5em]
{\footnotesize $^\dagger$Derived from Table 4 of Sprague et al.~\cite{sprague2016variation}; range is exact four-category agreement for consecutive exams interpreted by different radiologists (0.67) versus the same radiologist (0.80), not a confidence interval. Not directly comparable to model accuracy.}
\end{table*}

\subsection{External Validation on VinDr-Mammo}

Table~\ref{tab:vindr_auc_comparison} presents AUC comparison across score types on the VinDr-Mammo external validation set (N=\vitlVindrTriageN{} studies, \vitlVindrTriageNPos{} positive, \vitlVindrTriageNNeg{} negative).
All models are applied zero-shot with no VinDr-Mammo training data.
ViT-L achieves the highest triage AUC of \vitlVindrTriageAuc{} (95\% CI: \vitlVindrTriageAucLb{}--\vitlVindrTriageAucUb{}).
The detection head alone outperforms the triage head for ViT-L (\vitlVindrDetectionAuc{} vs.\ \vitlVindrTriageAuc{}), consistent with the pattern observed on the proprietary test set.
Model scaling is preserved, with ViT-L consistently outperforming ViT-B and ViT-S across all score types.

\begin{table*}[t]
\centering
\caption{AUC comparison across score types on the VinDr-Mammo external validation set (zero-shot, no VinDr training data). Triage uses the classification head alone; Detection uses the max-pooled heatmap score; Triage+Detection averages both; Ensemble combines ViT Triage+Detection with CNN Triage. N=\vitlVindrTriageN{} studies (\vitlVindrTriageNPos{} positive, \vitlVindrTriageNNeg{} negative). Ground truth is adjudicated BI-RADS assessment (4--5 vs.\ 1--2), not biopsy-confirmed.}
\label{tab:vindr_auc_comparison}
\begin{adjustbox}{max width=\textwidth}
\begin{tabular}{lrrrr}
\toprule
\textbf{Model} & \textbf{Triage} & \textbf{Detection} & \textbf{Triage+Det.} & \textbf{Ensemble} \\
\midrule
ViT-L & \vitlVindrTriageAuc{} [\vitlVindrTriageAucLb{}--\vitlVindrTriageAucUb{}] & \vitlVindrDetectionAuc{} [\vitlVindrDetectionAucLb{}--\vitlVindrDetectionAucUb{}] & \vitlVindrTriagedetectionAuc{} [\vitlVindrTriagedetectionAucLb{}--\vitlVindrTriagedetectionAucUb{}] & \vitlVindrEnsembleAuc{} [\vitlVindrEnsembleAucLb{}--\vitlVindrEnsembleAucUb{}] \\
ViT-B & \vitbVindrTriageAuc{} [\vitbVindrTriageAucLb{}--\vitbVindrTriageAucUb{}] & \vitbVindrDetectionAuc{} [\vitbVindrDetectionAucLb{}--\vitbVindrDetectionAucUb{}] & \vitbVindrTriagedetectionAuc{} [\vitbVindrTriagedetectionAucLb{}--\vitbVindrTriagedetectionAucUb{}] & \vitbVindrEnsembleAuc{} [\vitbVindrEnsembleAucLb{}--\vitbVindrEnsembleAucUb{}] \\
ViT-S & \vitsVindrTriageAuc{} [\vitsVindrTriageAucLb{}--\vitsVindrTriageAucUb{}] & \vitsVindrDetectionAuc{} [\vitsVindrDetectionAucLb{}--\vitsVindrDetectionAucUb{}] & \vitsVindrTriagedetectionAuc{} [\vitsVindrTriagedetectionAucLb{}--\vitsVindrTriagedetectionAucUb{}] & \vitsVindrEnsembleAuc{} [\vitsVindrEnsembleAucLb{}--\vitsVindrEnsembleAucUb{}] \\
CNN Baseline & \CnnVindrTriageAuc{} [\CnnVindrTriageAucLb{}--\CnnVindrTriageAucUb{}] & --- & --- & --- \\
\bottomrule
\end{tabular}
\end{adjustbox}
\end{table*}

Table~\ref{tab:vindr_published_comparison} contextualizes these results against published methods evaluated on VinDr-Mammo.
Despite operating in a zero-shot setting with no VinDr-Mammo training data, Aegis ViT-L achieves AUC of \vitlVindrTriagedetectionAuc{} using the Triage+Detection score, competitive with methods that were trained directly on VinDr-Mammo.

\begin{table*}[t]
\centering
\caption{Comparison with published methods on VinDr-Mammo. Direct comparison is limited by differences in positive class definition, evaluation level (image/breast/study), and training data. CV = cross-validation on VinDr-Mammo; Zero-shot = no VinDr-Mammo training data, thresholds from proprietary development set. Ground truth is adjudicated BI-RADS assessment, not biopsy-confirmed.}
\label{tab:vindr_published_comparison}
\begin{adjustbox}{max width=\textwidth}
\begin{tabular}{lrll}
\toprule
\textbf{Method} & \textbf{AUC} & \textbf{Training Data} & \textbf{Notes} \\
\midrule
ConvNeXt-Base PBC~\cite{petrini2025optimizing} & 0.851 & VinDr train & Image-level; BI-RADS 3--5 vs.\ 1--2 \\
MamT$^4$~\cite{ibragimov2024mamt4} & 0.840 & VinDr & Study-level; BI-RADS 4--5 vs.\ 1--2 \\
Dual-view~\cite{kebede2024dualview} & 0.830 & VinDr (CV) & Breast-level; BI-RADS $\geq$2 vs.\ 1 \\
AUCReshaping~\cite{bhat2023aucreshaping} & 0.770 & VinDr & Image-level; BI-RADS 4--5 vs.\ 1--2--3 \\
\midrule
Aegis ViT-L (T+D) & \vitlVindrTriagedetectionAuc{} & Zero-shot & Study-level; BI-RADS 4--5 vs.\ 1--2 \\
Aegis ViT-B (T+D) & \vitbVindrTriagedetectionAuc{} & Zero-shot & Study-level; BI-RADS 4--5 vs.\ 1--2 \\
Aegis ViT-S (T+D) & \vitsVindrTriagedetectionAuc{} & Zero-shot & Study-level; BI-RADS 4--5 vs.\ 1--2 \\
CNN Baseline & \CnnVindrTriageAuc{} & Zero-shot & Study-level; BI-RADS 4--5 vs.\ 1--2 \\
\bottomrule
\end{tabular}
\end{adjustbox}
\end{table*}

For breast density assessment on VinDr-Mammo, ViT-L achieves a binary (dense vs.\ non-dense) AUC of \vitlVindrDensityBinaryAuc{}
(95\% CI: \vitlVindrDensityBinaryAucLb{}--\vitlVindrDensityBinaryAucUb{}) and adjacent accuracy of \vitlVindrDensityAdjacentAccuracyMean{}
(95\% CI: \vitlVindrDensityAdjacentAccuracyCILower{}--\vitlVindrDensityAdjacentAccuracyCIUpper{}),
providing evidence that density assessment transfers to this external population.

\section{Discussion}

Aegis demonstrates strong performance on breast cancer triage, with the ViT-L variant achieving a triage AUC of \vitlTestfdaTriageAuc{} on the test set (or \vitlTestfdaEnsembleAuc{} with ensemble).
This represents meaningful improvement over the CNN baseline while maintaining clinically relevant sensitivity levels above 90\%.

\paragraph{Model Scaling}
We observe improvements with increased model capacity, with ViT-L achieving the highest performance across all metrics.
DeLong's test confirms that only ViT-L achieves statistically significant improvement over the CNN baseline (p=\delongVitlVsCnnPval{}),
while ViT-B and ViT-S show numerically higher AUC but do not reach significance (p=\delongVitbVsCnnPval{} and p=\delongVitsVsCnnPval{}, respectively).
This suggests that sufficient model capacity is necessary to fully leverage the self-supervised pre-training protocol
for learning discriminative representations from mammographic images.

\paragraph{Ensemble Benefits}
The combination of triage and detection heads provides complementary information that improves overall discrimination.
The ensemble approach achieves the highest AUC (\vitlTestfdaEnsembleAuc{}) while improving specificity without sacrificing sensitivity,
suggesting that the detection head captures lesion-level features that enhance study-level classification.

\paragraph{Density Assessment}
Breast density classification is inherently subjective, with substantial inter-reader variability among radiologists.
In the consecutive-exam analysis by Sprague et al., Table~4 implies four-category exact agreement of 67.4\% for exams interpreted by different radiologists and 80.1\% for exams interpreted by the same radiologist~\cite{sprague2016variation},
with non-exact assessments occurring almost exclusively between adjacent categories.
Our models achieve 4-class accuracy in the range of reported clinical-assessment agreement
while providing deterministic outputs for the same input image.
The high adjacent accuracy (\vitlDensityAdjacentAccuracyMean{}) for density classification indicates that the model's predictions are clinically reasonable even when not exactly correct,
as adjacent density categories often have overlapping radiographic features.

\paragraph{External Generalization}
External validation on VinDr-Mammo provides evidence that Aegis transfers across substantial domain shifts, including different patient populations (US vs.\ Vietnam), imaging manufacturers, and acquisition protocols.
The AUC reduction from the proprietary test set (ViT-L Triage: 0.945 vs.\ \vitlVindrTriageAuc{}) reflects both this domain shift and the difference in ground truth definition: biopsy-confirmed malignancy vs.\ adjudicated BI-RADS assessment without pathological confirmation.
Importantly, the model scaling trend is preserved on VinDr-Mammo (ViT-L $>$ ViT-B $>$ ViT-S), suggesting that larger model capacity provides more robust cross-domain representations.
Despite operating in a zero-shot setting with no VinDr-Mammo training data, Aegis achieves AUC competitive with published methods that were trained directly on VinDr-Mammo (Table~\ref{tab:vindr_published_comparison}), though direct comparison is limited by differences in evaluation methodology.

\paragraph{Limitations}
This study has several limitations. While VinDr-Mammo external validation partially addresses concerns about generalizability, that dataset uses adjudicated BI-RADS assessment rather than biopsy-confirmed ground truth; external validation on datasets with pathological confirmation remains needed.
The CNN baseline represents a specific architecture and training protocol;
comparisons with additional contemporary approaches would provide additional context for these results.

\paragraph{Computational Efficiency}
Table~\ref{tab:benchmarks} summarizes inference performance across hardware platforms for a typical study (batch of 4 mammogram images at 2048$\times$1536 resolution).
Models were exported using PyTorch's AOTInductor compiler to produce native shared libraries, which are loaded through a C++ bridge into a Rust application for inference.
This approach eliminates Python runtime dependencies and enables deployment as a single portable binary.
ROCm benchmarks utilize experimental Triton kernels for flash attention, available as of PyTorch 2.9.1.
The CNN baseline achieves the lowest latency on all platforms due to its simpler convolutional architecture; ROCm results are unavailable as no export exists for this model.
Among ViT models, latency scales approximately linearly with parameter count, with ViT-L requiring roughly twice the inference time of ViT-S.
CUDA acceleration provides 44--156$\times$ speedup over CPU for ViT models, making GPU deployment essential for practical throughput.
CPU inference remains viable only for the CNN model ($\sim$1.3~s per study) or in resource-constrained environments where longer latency is acceptable.
All models fit comfortably within consumer GPU memory (24~GB), enabling deployment on widely available hardware.
These results highlight a practical trade-off: ViT-L offers the highest diagnostic accuracy at approximately twice the latency and memory cost of ViT-S,
allowing deployment decisions to balance accuracy requirements against computational constraints.

\begin{table*}[t]
\centering
\caption{Inference benchmarks for a batch of 4 mammogram images at 2048$\times$1536 resolution using AOTInductor-compiled models. CUDA: RTX 3090 (CUDA 12.6); CPU: Threadripper 3960X; ROCm: Radeon Pro W7800 (ROCm 6.4). PyTorch 2.9.1.}
\label{tab:benchmarks}
\begin{adjustbox}{max width=\textwidth}
\begin{tabular}{lrcccc}
\toprule
\textbf{Model} & \textbf{Parameters} & \textbf{CUDA Latency} & \textbf{CUDA Memory} & \textbf{CPU Latency} & \textbf{ROCm Latency} \\
\midrule
ViT-L & 403M & 640 ms & 2565 MB & 100 s & 2781 ms \\
ViT-B & 114M & 415 ms & 1287 MB & 38 s & 1842 ms \\
ViT-S & 29M & 340 ms & 785 MB & 15 s & 1385 ms \\
CNN & 79M & 319 ms & 1007 MB & 1.3 s & --- \\
\bottomrule
\end{tabular}
\end{adjustbox}
\end{table*}

\paragraph{Future Work}
Several research directions could extend this work.
While VinDr-Mammo external validation supports cross-population transfer, evaluation on additional public datasets (e.g., INbreast) with biopsy-confirmed ground truth would further strengthen these claims.
Synthetic mammography data such as T-SYNTH~\cite{wiedeman2025tsynth} could augment training for rare lesion types and enable controlled evaluation studies.
Direct processing of 3D digital breast tomosynthesis volumes may improve detection of lesions obscured by overlapping tissue in 2D projections.
Model predictions could be leveraged for auto-annotation to refine dataset quality through pseudo-labeling and systematic identification of potential labeling errors.
Extended capabilities including breast arterial calcification analysis and integration of clinical risk factors could provide additional clinical value.
Incorporating prior examinations through longitudinal modeling would capture temporal changes for improved risk prediction.
Uncertainty quantification via Bayesian methods or conformal prediction would provide calibrated confidence estimates for clinical decision support.
Finally, prospective clinical validation studies measuring real-world impact on radiologist workflow and patient outcomes remain essential for deployment.

\section*{Acknowledgment}

This research was, in part, funded by the National Institutes of Health (NIH) Agreement No.\ 1OT2OD032581. The views and conclusions contained in this document are those of the authors and should not be interpreted as representing the official policies, either expressed or implied, of the NIH.

Chase Waggener thanks Jay Patel for his mentorship during the AIM-AHEAD Research Fellowship, during which much of the foundational work for this project was developed, and Rohan Senthilkumar for his assistance with annotation of training data.

\input{aegis_paper_2026.bbl}
\end{document}

%% file: tikz/macros.tex
\newcommand{\drawRCC}{
  \begin{tikzpicture}
    \fill[black] (0,0) rectangle (3, 4);
    \clip (0,0) rectangle (3, 4); 
    \fill[gray] (0, 0.2) arc[start angle=-90, end angle=90, radius=1.8] -- cycle;
    \draw[white, ultra thick] (0, 0.2) arc[start angle=-90, end angle=90, radius=1.8];
    \node[white, anchor=north east, inner sep=2, outer sep=2] at (3, 4) {RCC};
  \end{tikzpicture}
}

\newcommand{\drawRCCStacked}{
  \begin{tikzpicture}
    \foreach \i in {2,...,1} {
      \fill[gray] (0.2*\i, 0.2*\i) rectangle (3 + 0.2*\i, 4 + 0.2*\i);
      \draw[black] (0.2*\i, 0.2*\i) rectangle (3 + 0.2*\i, 4 + 0.2*\i);
    }
    \fill[black] (0,0) rectangle (3, 4);
    \clip (0,0) rectangle (3, 4); 
    \fill[gray] (0, 0.2) arc[start angle=-90, end angle=90, radius=1.8] -- cycle;
    \draw[white, ultra thick] (0, 0.2) arc[start angle=-90, end angle=90, radius=1.8];
    \node[white, anchor=north east, inner sep=2, outer sep=2] at (3, 4) {RCC};
  \end{tikzpicture}
}

\newcommand{\drawMaskedRCC}[6]{
  \pgfmathsetseed{#1}
  \begin{tikzpicture}
    \fill[black] (0,0) rectangle (3, 4);
    \clip (0,0) rectangle (3, 4); 
    \fill[gray] (0, 0.2) arc[start angle=-90, end angle=90, radius=1.8] -- cycle;
    \draw[white, ultra thick] (0, 0.2) arc[start angle=-90, end angle=90, radius=1.8];
    \node[white, anchor=north east] at (3, 4) {RCC};
    \foreach \x in {0, ..., \numexpr#5-1\relax} {
      \foreach \y in {0, ..., \numexpr#6-1\relax} {
        \pgfmathsetmacro{\width}{3/#5}
        \pgfmathsetmacro{\height}{4/#6}
        \draw[white, very thin] (\x*\width, \y*\height) rectangle (\x*\width+\width, \y*\height+\height);
        \pgfmathsetmacro{\rand}{random}
        \ifdim \rand pt < \dimexpr#2 pt\relax
          \fill[#3, opacity=#4] (\x*\width, \y*\height) rectangle (\x*\width+\width, \y*\height+\height);
        \else
          \fill[black] (\x*\width, \y*\height) rectangle (\x*\width+\width, \y*\height+\height);
        \fi
      }
    }
  \end{tikzpicture}
}

\newcommand{\drawMaskedRCCStacked}[6]{
  \pgfmathsetseed{#1}
  \begin{tikzpicture}
    \foreach \i in {2,...,1} {
      \fill[gray] (0.2*\i, 0.2*\i) rectangle (3 + 0.2*\i, 4 + 0.2*\i);
      \draw[black] (0.2*\i, 0.2*\i) rectangle (3 + 0.2*\i, 4 + 0.2*\i);
    }
    \fill[black] (0,0) rectangle (3, 4);
    \clip (0,0) rectangle (3, 4); 
    \fill[gray] (0, 0.2) arc[start angle=-90, end angle=90, radius=1.8] -- cycle;
    \draw[white, ultra thick] (0, 0.2) arc[start angle=-90, end angle=90, radius=1.8];
    \node[white, anchor=north east] at (3, 4) {RCC};
    \foreach \x in {0, ..., \numexpr#5-1\relax} {
      \foreach \y in {0, ..., \numexpr#6-1\relax} {
        \pgfmathsetmacro{\width}{3/#5}
        \pgfmathsetmacro{\height}{4/#6}
        \draw[white, very thin] (\x*\width, \y*\height) rectangle (\x*\width+\width, \y*\height+\height);
        \pgfmathsetmacro{\rand}{random}
        \ifdim \rand pt < \dimexpr#2 pt\relax
          \fill[#3, opacity=#4] (\x*\width, \y*\height) rectangle (\x*\width+\width, \y*\height+\height);
        \else
          \fill[black] (\x*\width, \y*\height) rectangle (\x*\width+\width, \y*\height+\height);
        \fi
      }
    }
  \end{tikzpicture}
}

\newcommand{\drawMLP}{
  \begin{tikzpicture}
    \foreach \i in {1, 2} {
      \node[circle, draw=black, fill=white, minimum size=10pt, inner sep=0pt] (input\i) at (0, -\i-0.5) {};
    }

    \foreach \i in {1, 2, 3} {
      \node[circle, draw=black, fill=white, minimum size=10pt, inner sep=0pt] (hidden\i) at (1, -\i) {};
    }

    \foreach \i in {1, 2} {
      \node[circle, draw=black, fill=white, minimum size=10pt, inner sep=0pt] (output\i) at (2, -\i-0.5) {};
    }

    \foreach \i in {1, 2} {
      \foreach \j in {1, 2, 3} {
        \draw[->, black, >=stealth] (input\i) -- (hidden\j);
      }
    }

    \foreach \i in {1, 2, 3} {
      \foreach \j in {1, 2} {
        \draw[->, black, >=stealth] (hidden\i) -- (output\j);
      }
    }
  \end{tikzpicture}
}

\newcommand{\drawFeatures}[3]{
  \begin{tikzpicture}
  \pgfmathsetseed{#1}
  \foreach \i in {0, ..., #2} {
    \node at (\i*0.8, 0) {
      $\color{#3}\scriptsize\begin{bmatrix}
        \pgfmathsetmacro{\rand}{rnd*9.9}
        \pgfmathprintnumber[fixed, precision=1]{\rand} \\
        \pgfmathsetmacro{\rand}{rnd*9.9}
        \pgfmathprintnumber[fixed, precision=1]{\rand} \\
        \vdots \\
        \pgfmathsetmacro{\rand}{rnd*9.9}
        \pgfmathprintnumber[fixed, precision=1]{\rand} \\
        \pgfmathsetmacro{\rand}{rnd*9.9}
        \pgfmathprintnumber[fixed, precision=1]{\rand}
      \end{bmatrix}$
    };
  }
  \end{tikzpicture}
}

\newcommand{\drawPosEnc}{
  \begin{tikzpicture}
  \node[circle, draw=black, minimum size=6cm, inner sep=1cm] (circle) at (0, 0) {};
  \draw[black] (-3, 0) sin (-1.5, -1) cos (0, 0) sin (1.5, 1) cos (3, 0);
  \end{tikzpicture}
}

\newcommand{\drawVector}[4]{
  \begin{tikzpicture}
    \foreach \i in {0,...,\numexpr#1-1\relax} {
      \ifthenelse{\equal{#2}{h}}{
        \fill[#3] (\i,0) rectangle (\i+1,1);
        \draw[#4] (\i,0) rectangle (\i+1,1);
      }{
        \fill[#3] (0,-\i*1) rectangle (1,-\i*1-1);
        \draw[#4] (0,-\i*1) rectangle (1,-\i*1-1);
      }
    }
  \end{tikzpicture}
}

\newcommand{\drawLayerBlock}[2]{
  \begin{tikzpicture}
    \tikzset{
        curved box/.style={
            draw,
            rounded corners=5pt,
            minimum width=1cm,
            minimum height=3cm,
            align=center,
            fill=#2
        }
    }
    \node[curved box, name=block] at (0,0) {\rotatebox{-90}{#1}};
  \end{tikzpicture}
}

\newcommand{\drawEncoderBlock}{
  \begin{tikzpicture}
    \node[name=sa] at (0, 0){\drawLayerBlock{Self Attention}{red!20}};
    \node[name=mlp] at (1.5, 0){\drawLayerBlock{SwiGLU MLP}{purple!20}};
    \draw[->, black, >=stealth] (sa) -- (mlp);

    \tikzset{
        curved box/.style={
            draw,
            rounded corners=5pt,
            minimum width=1cm,
            minimum height=3cm,
            align=center,
        }
    }
    \node[curved box, draw, fit=(sa) (mlp), inner sep=4pt] {};
  \end{tikzpicture}
}

\newcommand{\drawAddNode}{
  \begin{tikzpicture}
    \node[circle, draw, minimum size=10pt] (plus) at (0,0) {$+$};
  \end{tikzpicture}
}

\newcommand{\drawDetailedEncoderBlock}{
  \begin{tikzpicture}
    \node[outer sep=1pt, inner sep=0pt] (start) {};

    \def\blockspacing{0.4cm}

    \node[name=norm1, right=\blockspacing of start]{\drawLayerBlock{RMS Norm}{green!20}};
    \node[name=sa, right=\blockspacing of norm1]{\drawLayerBlock{Self Attention}{red!20}};
    \node[name=resid1, scale=0.75, outer sep=1pt, inner sep=1pt, right=\blockspacing of sa]{\drawAddNode};
    \draw[->, black, >=stealth] (start) -- (norm1);
    \draw[->, black, >=stealth] (norm1) -- (sa);
    \draw[->, black, >=stealth] (sa) -- (resid1);
    \draw[->, black, >=stealth] (start) 
        |- (norm1.north west) 
        -| (resid1.north west);

    \node[name=norm2, right=\blockspacing of resid1]{\drawLayerBlock{RMS Norm}{green!20}};
    \node[name=mlp, right=\blockspacing of norm2]{\drawLayerBlock{SwiGLU MLP}{purple!20}};
    \node[name=resid2, scale=0.75, outer sep=1pt, inner sep=1pt, right=\blockspacing of mlp]{\drawAddNode};
    \draw[->, black, >=stealth] (resid1) -- (norm2);
    \draw[->, black, >=stealth] (norm2) -- (mlp);
    \draw[->, black, >=stealth] (mlp) -- (resid2);
    \draw[->, black, >=stealth] (resid1.north east) 
        |- (norm2.north west) 
        -| (resid2);

    \tikzset{
        curved box/.style={
            draw,
            rounded corners=5pt,
            minimum width=1cm,
            minimum height=4cm,
            align=center,
            line width=1pt
        }
    }
    \node[curved box, draw, fit= (start) (norm1) (sa) (norm2) (mlp) (resid2), inner sep=4pt] {};
  \end{tikzpicture}
}

\newcommand{\drawDecoderBlock}{
  \begin{tikzpicture}
    \node[name=sa] at (0, 0){\drawLayerBlock{Self Attention}{red!20}};
    \node[name=ca] at (1.5, 0){\drawLayerBlock{Cross Attention}{green!20}};
    \node[name=mlp] at (3, 0){\drawLayerBlock{SwiGLU MLP}{purple!20}};
    \draw[->, black, >=stealth] (sa) -- (ca);
    \draw[->, black, >=stealth] (ca) -- (mlp);

    \tikzset{
        curved box/.style={
            draw,
            rounded corners=5pt,
            minimum width=1cm,
            minimum height=3cm,
            align=center,
        }
    }
    \node[curved box, draw, fit=(sa) (ca) (mlp), inner sep=4pt] {};
  \end{tikzpicture}
}

\newcommand{\drawConv}[3]{
  \begin{tikzpicture}
    \foreach \i in {0,...,#3} {
      \begin{scope}[yshift=\i*5pt]
        \foreach \x in {0,...,\numexpr#1-1\relax} {
          \foreach \y in {0,...,\numexpr#1-1\relax} {
            \draw[fill=#2] (\x*0.2,\y*0.2) rectangle ++(0.2,0.2);
          }
        }
      \end{scope}
    }
  \end{tikzpicture}
}

\newcommand{\drawDetailedDecoderBlock}[1]{
  \begin{tikzpicture}
    \node[outer sep=1pt, inner sep=0pt] (start) {};

    \def\blockspacing{0.4cm}

    \ifthenelse{\equal{#1}{t}}{
      \node[name=norm1, right=\blockspacing of start]{\drawLayerBlock{Layer Norm}{green!20}};
      \node[name=sa, right=\blockspacing of norm1]{\drawLayerBlock{Self Attention}{red!20}};
      \node[name=scale_sa, right=\blockspacing of sa]{\drawLayerBlock{Layer Scale}{gray!20}};
      \node[name=resid1, scale=0.75, outer sep=1pt, inner sep=1pt, right=\blockspacing of scale_sa]{\drawAddNode};
      \draw[->, black, >=stealth] (start) -- (norm1);
      \draw[->, black, >=stealth] (norm1) -- (sa);
      \draw[->, black, >=stealth] (sa) -- (scale_sa);
      \draw[->, black, >=stealth] (scale_sa) -- (resid1);
      \draw[->, black, >=stealth] (start) 
          |- (norm1.north west) 
          -| (resid1.north west);
    }{
      \node[name=resid1] at (start) {};
    }

    \node[name=norm2, right=\blockspacing of resid1]{\drawLayerBlock{Layer Norm}{green!20}};
    \node[name=ca, right=\blockspacing of norm2]{\drawLayerBlock{Cross Attention}{blue!20}};
    \node[name=scale_ca, right=\blockspacing of ca]{\drawLayerBlock{Layer Scale}{gray!20}};
    \node[name=resid2, scale=0.75, outer sep=1pt, inner sep=1pt, right=\blockspacing of scale_ca]{\drawAddNode};
    \draw[->, black, >=stealth] (resid1) -- (norm2);
    \draw[->, black, >=stealth] (norm2) -- (ca);
    \draw[->, black, >=stealth] (ca) -- (scale_ca);
    \draw[->, black, >=stealth] (scale_ca) -- (resid2);
    \ifthenelse{\equal{#1}{t}}{
      \draw[->, black, >=stealth] (resid1.north east) 
          |- (norm2.north west) 
          -| (resid2);
    }{
      \draw[->, black, >=stealth] (start.north east)
          |- (norm2.north west)
          -| (resid2);
    }

    \node[name=norm3, right=\blockspacing of resid2]{\drawLayerBlock{RMS Norm}{green!20}};
    \node[name=mlp, right=\blockspacing of norm3]{\drawLayerBlock{SwiGLU MLP}{purple!20}};
    \node[name=resid3, scale=0.75, outer sep=1pt, inner sep=1pt, right=\blockspacing of mlp]{\drawAddNode};
    \draw[->, black, >=stealth] (resid2) -- (norm3);
    \draw[->, black, >=stealth] (norm3) -- (mlp);
    \draw[->, black, >=stealth] (mlp) -- (resid3);
    \draw[->, black, >=stealth] (resid2.north east) 
        |- (norm3.north west) 
        -| (resid3);

    \tikzset{
        curved box/.style={
            draw,
            rounded corners=5pt,
            minimum width=1cm,
            minimum height=4cm,
            align=center,
            line width=1pt
        }
    }
    \node[curved box, draw, fit= (start) (norm2) (ca) (norm3) (mlp) (resid3), inner sep=4pt] {};

    \coordinate (ca_north) at (ca.north);
    \coordinate (ca_south) at (ca.south);
    \coordinate (ca_east) at (ca.east); 
    \coordinate (ca_west) at (ca.west);
  \end{tikzpicture}
}

%% file: tikz/jepa.tex
\tikzset{every picture/.style={/utils/exec={\sffamily}}}

\resizebox{\textwidth}{!}{
\begin{tikzpicture}

  \node[name=rcc] at (0, 0){\drawRCC};
  \node[above=0.5cm of rcc.north, anchor=center] {Mammogram};

  \node[name=teacher] at (4, 0){\drawMLP};
  \node[above=0.5cm of teacher.north, anchor=center] {Teacher Network};
  \draw[->, thick, black, >=stealth, line width=1.5pt] (rcc) -- (teacher);

  \node[name=masked_rcc] at (0, -6){\drawMaskedRCC{3}{0.5}{blue}{0.0}{3}{4}};
  \node[below=0.5cm of masked_rcc.south, anchor=center] {Masked Mammogram};
  \draw[->, thick, black, >=stealth, line width=1.5pt, align=center] 
    (rcc) 
    to
    node[midway, right] 
    {Augment (noise, MixUp)\\ Mask ($p=0.5$)} 
    (masked_rcc);

  \node[name=student] at (4, -6){\drawMLP};
  \node[below=0.5cm of student.south, anchor=center] {Student Network};
  \draw[->, thick, black, >=stealth, line width=1.5pt] (masked_rcc) -- (student);

  \draw[->, thick, red, dashed, >=stealth, line width=1.5pt] 
    (student) 
    to[out=75, in=-75]
    node[midway, right] 
    {EMA Update}
    (teacher);

  \node[name=student_features] at (8, -6){\drawMaskedRCC{3}{0.5}{blue}{0.25}{3}{4}};
  \node[below=0.5cm of student_features.south, anchor=center] {Student Features};
  \draw[->, thick, black, >=stealth, line width=1.5pt] (student) -- (student_features);

  \node[name=teacher_features] at (8, 0){\drawMaskedRCC{5}{1.0}{red}{0.25}{3}{4}};
  \node[above=0.5cm of teacher_features.north, anchor=center] {Teacher Features};
  \draw[->, thick, black, >=stealth, line width=1.5pt] 
    (teacher) 
    to
    node[midway, above] {MixUp}
    (teacher_features);

  \node[name=masked_teacher_features] at (14, 0){\drawMaskedRCC{8}{0.25}{red}{0.25}{3}{4}};
  \node[above=0.5cm of masked_teacher_features.north, anchor=center] {Selected Teacher Features};
  \draw[->, thick, black, >=stealth, line width=1.5pt] 
    (teacher_features) 
    to
    node[midway, above] 
    {Subsample}
    node[midway, below] 
    {$p=0.25$}
    (masked_teacher_features);

  \node[name=predictor] at (14, -6){\drawMLP};
  \node[below=0.5cm of predictor.south, anchor=center] {Predictor Network};
  \node[below=0.9cm of predictor.south, anchor=center, font=\scriptsize, text=purple] {(2× passes: visual + CLS tokens)};
  \draw[->, thick, black, >=stealth, line width=1.5pt] 
    (student_features) 
    to
    node[pos=0.9, below] {\scriptsize KV}
    (predictor);

  \draw[->, thick, black, >=stealth, line width=1.5pt] 
    (masked_teacher_features.south) 
    to[out=270, in=90] 
    node[midway, left] 
    {Position Encodings} 
    node[pos=0.9, right] {\scriptsize Q}
    (predictor.north);

  \node[name=predictions] at (19, -6){\drawMaskedRCC{8}{0.25}{green}{0.25}{3}{4}};
  \node[below=0.5cm of predictions.south, anchor=center] {Predictions};
  \draw[->, thick, black, >=stealth, line width=1.5pt] (predictor) -- (predictions);

  \node[name=loss] at (19, 0){\Large $\mathcal{L}(x, \hat{x})$};
  \node[above=0.5cm of loss.north, anchor=center] {Mean Squared Error Loss};

  \draw[->, thick, black, >=stealth, line width=1.5pt] (masked_teacher_features) to[out=0, in=180] (loss);
  \draw[->, thick, black, >=stealth, line width=1.5pt] (predictions.north) to[out=90, in=270] (loss);

\end{tikzpicture}
}

%% file: tikz/finetune.tex
\tikzset{every picture/.style={/utils/exec={\sffamily}}}

\resizebox{\textwidth}{!}{
\begin{tikzpicture}

  \node[name=input] at (0, 0){\drawRCC};
  \node[above=0.5cm of input.north, anchor=center] {Mammogram};

  \node[name=vit] at (5, 0){\drawEncoderBlock};
  \node[above=0.5cm of vit.north, anchor=center] {ViT Backbone};
  \draw[->, thick, black, >=stealth, line width=1.5pt] (input) -- (vit);

  \node[name=visual_tokens] at (10, 2){\drawMaskedRCC{42}{1.0}{blue}{0.5}{4}{5}};
  \node[above=0.3cm of visual_tokens.north, anchor=center, font=\small] {Visual Tokens};
  \draw[->, thick, black, >=stealth, line width=1.5pt] 
    (vit.east) 
    to[out=0, in=180] 
    (visual_tokens.west);

  \node[name=cls_tokens] at (10, -3) {
    \begin{tikzpicture}
      \foreach \i/\col in {0/red, 1/green, 2/orange, 3/purple} {
        \fill[\col!60] (0, -\i*0.8) rectangle (0.7, -\i*0.8-0.7);
        \draw[black, thick] (0, -\i*0.8) rectangle (0.7, -\i*0.8-0.7);
        \node[font=\tiny, inner sep=2pt] at (0.35, -\i*0.8-0.35) {CLS\i};
      }
    \end{tikzpicture}
  };
  \node[below=0.3cm of cls_tokens.south, anchor=center, font=\small] {CLS Tokens (4)};

  \coordinate (cls0) at (10.35, -1.8);
  \coordinate (cls1) at (10.35, -2.6);
  \coordinate (cls2) at (10.35, -3.4);
  \coordinate (cls3) at (10.35, -4.2);
  
  \draw[->, thick, black, >=stealth, line width=1.5pt] 
    (vit.east) 
    to[out=0, in=180] 
    (cls_tokens.west);

  \node[name=heatmap_head, rectangle, draw, fill=blue!20, minimum width=1.5cm, minimum height=0.8cm] at (15, 3.0) {Head};
  \node[above=0.1cm of heatmap_head.north, anchor=center, font=\tiny] {Detection Head};
  \node[name=heatmap_out, rectangle, draw, fill=blue!10, minimum width=4cm, minimum height=0.6cm] at (18.5, 3.0) {Lesion Heatmap};
  \draw[->, thick, blue!60, >=stealth, line width=1.5pt] 
    (visual_tokens.east) 
    to[out=0, in=180] 
    (heatmap_head.west);
  \draw[->, thick, black, >=stealth, line width=1.2pt] (heatmap_head) -- (heatmap_out);

  \node[name=tissue_head, rectangle, draw, fill=cyan!20, minimum width=1.5cm, minimum height=0.8cm] at (15, 1.0) {Head};
  \node[above=0.1cm of tissue_head.north, anchor=center, font=\tiny] {Tissue Head};
  \node[name=tissue_out] at (18.5, 1.0) {
    \begin{tikzpicture}
      \node[rectangle, draw, fill=cyan!10, minimum width=2.2cm, minimum height=0.4cm, font=\tiny] at (0, 0.5) {Pectoral Muscle};
      \node[rectangle, draw, fill=cyan!10, minimum width=2.2cm, minimum height=0.4cm, font=\tiny] at (0, 0) {Breast Tissue};
      \node[rectangle, draw, fill=cyan!10, minimum width=2.2cm, minimum height=0.4cm, font=\tiny] at (0, -0.5) {Background};
    \end{tikzpicture}
  };
  \draw[->, thick, cyan!60, >=stealth, line width=1.5pt]
    (visual_tokens.east)
    to[out=0, in=180]
    (tissue_head.west);
  \draw[->, thick, black, >=stealth, line width=1.2pt] (tissue_head) -- (tissue_out);

  \node[name=triage_head, rectangle, draw, fill=red!20, minimum width=1.5cm, minimum height=0.8cm] at (15, -1.5) {Head};
  \node[above=0.1cm of triage_head.north, anchor=center, font=\tiny] {Triage Head};
  \node[name=triage_out, rectangle, draw, fill=red!10, minimum width=3cm, minimum height=0.6cm] at (18, -1.5) {Malignancy};
  \draw[->, thick, red!60, >=stealth, line width=1.2pt] 
    (cls0) 
    to[out=0, in=180] 
    node[pos=0.8, above, font=\tiny] {CLS0}
    (triage_head.west);
  \draw[->, thick, black, >=stealth, line width=1.2pt] (triage_head) -- (triage_out);

  \node[name=view_head, rectangle, draw, fill=green!20, minimum width=1.5cm, minimum height=0.8cm] at (15, -2.8) {Head};
  \node[above=0.1cm of view_head.north, anchor=center, font=\tiny] {View Head};
  \node[name=view_out] at (18.5, -2.8) {
    \begin{tikzpicture}
      \node[rectangle, draw, fill=green!10, minimum width=1.2cm, minimum height=0.4cm, font=\tiny] at (0, 0.5) {MLO};
      \node[rectangle, draw, fill=green!10, minimum width=1.2cm, minimum height=0.4cm, font=\tiny] at (0, 0) {CC};
      \node[rectangle, draw, fill=green!10, minimum width=1.2cm, minimum height=0.4cm, font=\tiny] at (0, -0.5) {Spot/Mag};
    \end{tikzpicture}
  };
  \draw[->, thick, green!60, >=stealth, line width=1.2pt] 
    (cls1) 
    to[out=0, in=180] 
    node[pos=0.8, above, font=\tiny] {CLS1}
    (view_head.west);
  \draw[->, thick, black, >=stealth, line width=1.2pt] (view_head) -- (view_out);

  \node[name=implants_head, rectangle, draw, fill=orange!20, minimum width=1.5cm, minimum height=0.8cm] at (15, -4.8) {Head};
  \node[above=0.1cm of implants_head.north, anchor=center, font=\tiny] {Implants Head};
  \node[name=implants_out, rectangle, draw, fill=orange!10, minimum width=3cm, minimum height=0.6cm] at (18, -4.8) {Implants Present};
  \draw[->, thick, orange!80, >=stealth, line width=1.2pt] 
    (cls2) 
    to[out=0, in=180] 
    node[pos=0.8, above, font=\tiny] {CLS2}
    (implants_head.west);
  \draw[->, thick, black, >=stealth, line width=1.2pt] (implants_head) -- (implants_out);

  \node[name=density_head, rectangle, draw, fill=purple!20, minimum width=1.5cm, minimum height=0.8cm] at (15, -6.1) {Head};
  \node[above=0.1cm of density_head.north, anchor=center, font=\tiny] {Density Head};
  \node[name=density_out, rectangle, draw, fill=purple!10, minimum width=3cm, minimum height=0.6cm] at (18, -6.1) {Density Score};
  \draw[->, thick, purple!80, >=stealth, line width=1.2pt] 
    (cls3) 
    to[out=0, in=180] 
    node[pos=0.8, above, font=\tiny] {CLS3}
    (density_head.west);
  \draw[->, thick, black, >=stealth, line width=1.2pt] (density_head) -- (density_out);

\end{tikzpicture}
}

%% file: tables/data_variables.tex
\newcommand{\clinicalSitesPretrain}{14}
\newcommand{\totalStudiesPretrain}{71,103}
\newcommand{\totalPatientsPretrain}{31,110}
\newcommand{\totalImagesPretrain}{307,932}

\newcommand{\clinicalSitesTrain}{13}
\newcommand{\totalStudiesTrain}{68,721}
\newcommand{\totalPatientsTrain}{29,611}
\newcommand{\totalImagesTrain}{243,315}
\newcommand{\totalStudiesTrainWithMalignancy}{20,041}
\newcommand{\totalPatientsTrainWithMalignancy}{9,702}

\newcommand{\clinicalSitesDev}{2}
\newcommand{\totalStudiesDev}{2,107}
\newcommand{\totalPatientsDev}{1,584}
\newcommand{\totalImagesDev}{12,509}

\newcommand{\validStudiesDev}{2,095}

\newcommand{\clinicalSitesTest}{12}
\newcommand{\totalStudiesTest}{2,262}
\newcommand{\totalPatientsTest}{2,262}
\newcommand{\totalImagesTest}{18,096}
\newcommand{\selectedStudiesTest}{2,262}
\newcommand{\selectedStudiesTestMalignant}{1,054}
\newcommand{\selectedStudiesTestBenign}{1,208}

\newcommand{\curatedStudiesTest}{785}
\newcommand{\curatedStudiesTestMalignant}{394}
\newcommand{\curatedStudiesTestBenign}{391}

\newcommand{\totalStudies}{75,472}
\newcommand{\totalPatients}{34,956}
\newcommand{\totalImages}{338,537}

%% file: tables/data_partition_summary.tex
\begin{table}[tbp]
\centering
\begin{tabular}{lrrrr}
\toprule
Partition & Clinical Sites & Patients & Studies & Images \\
\midrule
Pretrain & 14 & 31,110 & 71,103 & 307,932 \\
Train & 13 & 29,611 & 68,721 & 243,315 \\
Dev & 2 & 1,584 & 2,107 & 12,509 \\
Test & 12 & 2,262 & 2,262 & 18,096 \\
\bottomrule
\end{tabular}
\caption{Summary of data partitions showing the number of clinical sites, patients, studies, and images in each partition. Note: Pretrain is a superset containing all Train data.}
\label{tab:data_partition_summary}
\end{table}

%% file: results/combined/variables.tex
\providecommand{\vitlDevTriageN}{2095}
\providecommand{\vitlDevTriageNPos}{626}
\providecommand{\vitlDevTriageNNeg}{1469}
\providecommand{\vitlDevTriageAuc}{0.904}
\providecommand{\vitlDevTriageAucLb}{0.883}
\providecommand{\vitlDevTriageAucUb}{0.922}
\providecommand{\vitlDevTriageSens}{80.7}
\providecommand{\vitlDevTriageSensLb}{77.6}
\providecommand{\vitlDevTriageSensUb}{83.7}
\providecommand{\vitlDevTriageSpec}{81.0}
\providecommand{\vitlDevTriageSpecLb}{79.0}
\providecommand{\vitlDevTriageSpecUb}{83.0}
\providecommand{\vitlDevTriagePpv}{64.4}
\providecommand{\vitlDevTriageNpv}{90.8}

\providecommand{\vitbDevTriageN}{2095}
\providecommand{\vitbDevTriageNPos}{626}
\providecommand{\vitbDevTriageNNeg}{1469}
\providecommand{\vitbDevTriageAuc}{0.879}
\providecommand{\vitbDevTriageAucLb}{0.855}
\providecommand{\vitbDevTriageAucUb}{0.901}
\providecommand{\vitbDevTriageSens}{78.9}
\providecommand{\vitbDevTriageSensLb}{75.7}
\providecommand{\vitbDevTriageSensUb}{82.1}
\providecommand{\vitbDevTriageSpec}{78.6}
\providecommand{\vitbDevTriageSpecLb}{76.3}
\providecommand{\vitbDevTriageSpecUb}{80.7}
\providecommand{\vitbDevTriagePpv}{61.1}
\providecommand{\vitbDevTriageNpv}{89.7}

\providecommand{\vitsDevTriageN}{2095}
\providecommand{\vitsDevTriageNPos}{626}
\providecommand{\vitsDevTriageNNeg}{1469}
\providecommand{\vitsDevTriageAuc}{0.874}
\providecommand{\vitsDevTriageAucLb}{0.852}
\providecommand{\vitsDevTriageAucUb}{0.895}
\providecommand{\vitsDevTriageSens}{78.4}
\providecommand{\vitsDevTriageSensLb}{75.4}
\providecommand{\vitsDevTriageSensUb}{81.8}
\providecommand{\vitsDevTriageSpec}{78.2}
\providecommand{\vitsDevTriageSpecLb}{76.0}
\providecommand{\vitsDevTriageSpecUb}{80.3}
\providecommand{\vitsDevTriagePpv}{60.5}
\providecommand{\vitsDevTriageNpv}{89.5}

\providecommand{\CnnDevTriageN}{2095}
\providecommand{\CnnDevTriageNPos}{626}
\providecommand{\CnnDevTriageNNeg}{1469}
\providecommand{\CnnDevTriageAuc}{0.876}
\providecommand{\CnnDevTriageAucLb}{0.852}
\providecommand{\CnnDevTriageAucUb}{0.897}
\providecommand{\CnnDevTriageSens}{79.7}
\providecommand{\CnnDevTriageSensLb}{76.5}
\providecommand{\CnnDevTriageSensUb}{82.7}
\providecommand{\CnnDevTriageSpec}{79.9}
\providecommand{\CnnDevTriageSpecLb}{77.8}
\providecommand{\CnnDevTriageSpecUb}{81.8}
\providecommand{\CnnDevTriagePpv}{62.8}
\providecommand{\CnnDevTriageNpv}{90.2}

\providecommand{\vitlTestfdaTriageN}{785}
\providecommand{\vitlTestfdaTriageNPos}{394}
\providecommand{\vitlTestfdaTriageNNeg}{391}
\providecommand{\vitlTestfdaTriageAuc}{0.945}
\providecommand{\vitlTestfdaTriageAucLb}{0.923}
\providecommand{\vitlTestfdaTriageAucUb}{0.966}
\providecommand{\vitlTestfdaTriageSens}{92.9}
\providecommand{\vitlTestfdaTriageSensLb}{90.4}
\providecommand{\vitlTestfdaTriageSensUb}{95.4}
\providecommand{\vitlTestfdaTriageSpec}{74.4}
\providecommand{\vitlTestfdaTriageSpecLb}{70.3}
\providecommand{\vitlTestfdaTriageSpecUb}{79.0}
\providecommand{\vitlTestfdaTriagePpv}{78.5}
\providecommand{\vitlTestfdaTriageNpv}{91.2}

\providecommand{\vitbTestfdaTriageN}{785}
\providecommand{\vitbTestfdaTriageNPos}{394}
\providecommand{\vitbTestfdaTriageNNeg}{391}
\providecommand{\vitbTestfdaTriageAuc}{0.924}
\providecommand{\vitbTestfdaTriageAucLb}{0.898}
\providecommand{\vitbTestfdaTriageAucUb}{0.949}
\providecommand{\vitbTestfdaTriageSens}{92.9}
\providecommand{\vitbTestfdaTriageSensLb}{90.6}
\providecommand{\vitbTestfdaTriageSensUb}{95.4}
\providecommand{\vitbTestfdaTriageSpec}{62.4}
\providecommand{\vitbTestfdaTriageSpecLb}{57.5}
\providecommand{\vitbTestfdaTriageSpecUb}{67.0}
\providecommand{\vitbTestfdaTriagePpv}{71.3}
\providecommand{\vitbTestfdaTriageNpv}{89.7}

\providecommand{\vitsTestfdaTriageN}{785}
\providecommand{\vitsTestfdaTriageNPos}{394}
\providecommand{\vitsTestfdaTriageNNeg}{391}
\providecommand{\vitsTestfdaTriageAuc}{0.935}
\providecommand{\vitsTestfdaTriageAucLb}{0.911}
\providecommand{\vitsTestfdaTriageAucUb}{0.957}
\providecommand{\vitsTestfdaTriageSens}{92.9}
\providecommand{\vitsTestfdaTriageSensLb}{90.4}
\providecommand{\vitsTestfdaTriageSensUb}{95.2}
\providecommand{\vitsTestfdaTriageSpec}{73.4}
\providecommand{\vitsTestfdaTriageSpecLb}{69.1}
\providecommand{\vitsTestfdaTriageSpecUb}{77.5}
\providecommand{\vitsTestfdaTriagePpv}{77.9}
\providecommand{\vitsTestfdaTriageNpv}{91.1}

\providecommand{\CnnTestfdaTriageN}{785}
\providecommand{\CnnTestfdaTriageNPos}{394}
\providecommand{\CnnTestfdaTriageNNeg}{391}
\providecommand{\CnnTestfdaTriageAuc}{0.925}
\providecommand{\CnnTestfdaTriageAucLb}{0.898}
\providecommand{\CnnTestfdaTriageAucUb}{0.948}
\providecommand{\CnnTestfdaTriageSens}{86.0}
\providecommand{\CnnTestfdaTriageSensLb}{82.7}
\providecommand{\CnnTestfdaTriageSensUb}{89.6}
\providecommand{\CnnTestfdaTriageSpec}{80.6}
\providecommand{\CnnTestfdaTriageSpecLb}{76.5}
\providecommand{\CnnTestfdaTriageSpecUb}{84.4}
\providecommand{\CnnTestfdaTriagePpv}{81.7}
\providecommand{\CnnTestfdaTriageNpv}{85.1}

\providecommand{\vitlDevDetectionN}{2095}
\providecommand{\vitlDevDetectionNPos}{626}
\providecommand{\vitlDevDetectionNNeg}{1469}
\providecommand{\vitlDevDetectionAuc}{0.905}
\providecommand{\vitlDevDetectionAucLb}{0.884}
\providecommand{\vitlDevDetectionAucUb}{0.923}
\providecommand{\vitlDevDetectionSens}{82.4}
\providecommand{\vitlDevDetectionSensLb}{79.4}
\providecommand{\vitlDevDetectionSensUb}{85.1}
\providecommand{\vitlDevDetectionSpec}{82.2}
\providecommand{\vitlDevDetectionSpecLb}{80.3}
\providecommand{\vitlDevDetectionSpecUb}{84.1}
\providecommand{\vitlDevDetectionPpv}{66.4}
\providecommand{\vitlDevDetectionNpv}{91.7}

\providecommand{\vitbDevDetectionN}{2095}
\providecommand{\vitbDevDetectionNPos}{626}
\providecommand{\vitbDevDetectionNNeg}{1469}
\providecommand{\vitbDevDetectionAuc}{0.884}
\providecommand{\vitbDevDetectionAucLb}{0.862}
\providecommand{\vitbDevDetectionAucUb}{0.905}
\providecommand{\vitbDevDetectionSens}{78.9}
\providecommand{\vitbDevDetectionSensLb}{75.7}
\providecommand{\vitbDevDetectionSensUb}{81.8}
\providecommand{\vitbDevDetectionSpec}{79.0}
\providecommand{\vitbDevDetectionSpecLb}{76.9}
\providecommand{\vitbDevDetectionSpecUb}{80.9}
\providecommand{\vitbDevDetectionPpv}{61.6}
\providecommand{\vitbDevDetectionNpv}{89.8}

\providecommand{\vitsDevDetectionN}{2095}
\providecommand{\vitsDevDetectionNPos}{626}
\providecommand{\vitsDevDetectionNNeg}{1469}
\providecommand{\vitsDevDetectionAuc}{0.872}
\providecommand{\vitsDevDetectionAucLb}{0.849}
\providecommand{\vitsDevDetectionAucUb}{0.893}
\providecommand{\vitsDevDetectionSens}{78.8}
\providecommand{\vitsDevDetectionSensLb}{75.4}
\providecommand{\vitsDevDetectionSensUb}{81.8}
\providecommand{\vitsDevDetectionSpec}{79.0}
\providecommand{\vitsDevDetectionSpecLb}{76.7}
\providecommand{\vitsDevDetectionSpecUb}{80.9}
\providecommand{\vitsDevDetectionPpv}{61.5}
\providecommand{\vitsDevDetectionNpv}{89.7}

\providecommand{\vitlTestfdaDetectionN}{785}
\providecommand{\vitlTestfdaDetectionNPos}{394}
\providecommand{\vitlTestfdaDetectionNNeg}{391}
\providecommand{\vitlTestfdaDetectionAuc}{0.942}
\providecommand{\vitlTestfdaDetectionAucLb}{0.918}
\providecommand{\vitlTestfdaDetectionAucUb}{0.963}
\providecommand{\vitlTestfdaDetectionSens}{92.4}
\providecommand{\vitlTestfdaDetectionSensLb}{89.8}
\providecommand{\vitlTestfdaDetectionSensUb}{94.9}
\providecommand{\vitlTestfdaDetectionSpec}{74.9}
\providecommand{\vitlTestfdaDetectionSpecLb}{70.3}
\providecommand{\vitlTestfdaDetectionSpecUb}{79.3}
\providecommand{\vitlTestfdaDetectionPpv}{78.8}
\providecommand{\vitlTestfdaDetectionNpv}{90.7}

\providecommand{\vitbTestfdaDetectionN}{785}
\providecommand{\vitbTestfdaDetectionNPos}{394}
\providecommand{\vitbTestfdaDetectionNNeg}{391}
\providecommand{\vitbTestfdaDetectionAuc}{0.927}
\providecommand{\vitbTestfdaDetectionAucLb}{0.901}
\providecommand{\vitbTestfdaDetectionAucUb}{0.951}
\providecommand{\vitbTestfdaDetectionSens}{91.1}
\providecommand{\vitbTestfdaDetectionSensLb}{88.3}
\providecommand{\vitbTestfdaDetectionSensUb}{93.9}
\providecommand{\vitbTestfdaDetectionSpec}{65.0}
\providecommand{\vitbTestfdaDetectionSpecLb}{60.4}
\providecommand{\vitbTestfdaDetectionSpecUb}{69.8}
\providecommand{\vitbTestfdaDetectionPpv}{72.4}
\providecommand{\vitbTestfdaDetectionNpv}{87.9}

\providecommand{\vitsTestfdaDetectionN}{785}
\providecommand{\vitsTestfdaDetectionNPos}{394}
\providecommand{\vitsTestfdaDetectionNNeg}{391}
\providecommand{\vitsTestfdaDetectionAuc}{0.928}
\providecommand{\vitsTestfdaDetectionAucLb}{0.904}
\providecommand{\vitsTestfdaDetectionAucUb}{0.951}
\providecommand{\vitsTestfdaDetectionSens}{89.6}
\providecommand{\vitsTestfdaDetectionSensLb}{86.8}
\providecommand{\vitsTestfdaDetectionSensUb}{92.6}
\providecommand{\vitsTestfdaDetectionSpec}{71.1}
\providecommand{\vitsTestfdaDetectionSpecLb}{66.5}
\providecommand{\vitsTestfdaDetectionSpecUb}{75.4}
\providecommand{\vitsTestfdaDetectionPpv}{75.8}
\providecommand{\vitsTestfdaDetectionNpv}{87.1}

\providecommand{\vitlDevTriagedetectionN}{2095}
\providecommand{\vitlDevTriagedetectionNPos}{626}
\providecommand{\vitlDevTriagedetectionNNeg}{1469}
\providecommand{\vitlDevTriagedetectionAuc}{0.911}
\providecommand{\vitlDevTriagedetectionAucLb}{0.891}
\providecommand{\vitlDevTriagedetectionAucUb}{0.929}
\providecommand{\vitlDevTriagedetectionSens}{82.4}
\providecommand{\vitlDevTriagedetectionSensLb}{79.4}
\providecommand{\vitlDevTriagedetectionSensUb}{85.5}
\providecommand{\vitlDevTriagedetectionSpec}{82.4}
\providecommand{\vitlDevTriagedetectionSpecLb}{80.3}
\providecommand{\vitlDevTriagedetectionSpecUb}{84.3}
\providecommand{\vitlDevTriagedetectionPpv}{66.6}
\providecommand{\vitlDevTriagedetectionNpv}{91.7}

\providecommand{\vitbDevTriagedetectionN}{2095}
\providecommand{\vitbDevTriagedetectionNPos}{626}
\providecommand{\vitbDevTriagedetectionNNeg}{1469}
\providecommand{\vitbDevTriagedetectionAuc}{0.890}
\providecommand{\vitbDevTriagedetectionAucLb}{0.867}
\providecommand{\vitbDevTriagedetectionAucUb}{0.910}
\providecommand{\vitbDevTriagedetectionSens}{79.9}
\providecommand{\vitbDevTriagedetectionSensLb}{76.7}
\providecommand{\vitbDevTriagedetectionSensUb}{82.9}
\providecommand{\vitbDevTriagedetectionSpec}{79.7}
\providecommand{\vitbDevTriagedetectionSpecLb}{77.5}
\providecommand{\vitbDevTriagedetectionSpecUb}{81.8}
\providecommand{\vitbDevTriagedetectionPpv}{62.7}
\providecommand{\vitbDevTriagedetectionNpv}{90.3}

\providecommand{\vitsDevTriagedetectionN}{2095}
\providecommand{\vitsDevTriagedetectionNPos}{626}
\providecommand{\vitsDevTriagedetectionNNeg}{1469}
\providecommand{\vitsDevTriagedetectionAuc}{0.882}
\providecommand{\vitsDevTriagedetectionAucLb}{0.859}
\providecommand{\vitsDevTriagedetectionAucUb}{0.901}
\providecommand{\vitsDevTriagedetectionSens}{79.9}
\providecommand{\vitsDevTriagedetectionSensLb}{76.7}
\providecommand{\vitsDevTriagedetectionSensUb}{82.7}
\providecommand{\vitsDevTriagedetectionSpec}{80.5}
\providecommand{\vitsDevTriagedetectionSpecLb}{78.4}
\providecommand{\vitsDevTriagedetectionSpecUb}{82.6}
\providecommand{\vitsDevTriagedetectionPpv}{63.5}
\providecommand{\vitsDevTriagedetectionNpv}{90.4}

\providecommand{\vitlTestfdaTriagedetectionN}{785}
\providecommand{\vitlTestfdaTriagedetectionNPos}{394}
\providecommand{\vitlTestfdaTriagedetectionNNeg}{391}
\providecommand{\vitlTestfdaTriagedetectionAuc}{0.949}
\providecommand{\vitlTestfdaTriagedetectionAucLb}{0.927}
\providecommand{\vitlTestfdaTriagedetectionAucUb}{0.969}
\providecommand{\vitlTestfdaTriagedetectionSens}{93.1}
\providecommand{\vitlTestfdaTriagedetectionSensLb}{90.6}
\providecommand{\vitlTestfdaTriagedetectionSensUb}{95.7}
\providecommand{\vitlTestfdaTriagedetectionSpec}{74.7}
\providecommand{\vitlTestfdaTriagedetectionSpecLb}{70.3}
\providecommand{\vitlTestfdaTriagedetectionSpecUb}{79.3}
\providecommand{\vitlTestfdaTriagedetectionPpv}{78.8}
\providecommand{\vitlTestfdaTriagedetectionNpv}{91.5}

\providecommand{\vitbTestfdaTriagedetectionN}{785}
\providecommand{\vitbTestfdaTriagedetectionNPos}{394}
\providecommand{\vitbTestfdaTriagedetectionNNeg}{391}
\providecommand{\vitbTestfdaTriagedetectionAuc}{0.933}
\providecommand{\vitbTestfdaTriagedetectionAucLb}{0.908}
\providecommand{\vitbTestfdaTriagedetectionAucUb}{0.956}
\providecommand{\vitbTestfdaTriagedetectionSens}{94.7}
\providecommand{\vitbTestfdaTriagedetectionSensLb}{92.1}
\providecommand{\vitbTestfdaTriagedetectionSensUb}{96.7}
\providecommand{\vitbTestfdaTriagedetectionSpec}{64.2}
\providecommand{\vitbTestfdaTriagedetectionSpecLb}{59.6}
\providecommand{\vitbTestfdaTriagedetectionSpecUb}{69.3}
\providecommand{\vitbTestfdaTriagedetectionPpv}{72.7}
\providecommand{\vitbTestfdaTriagedetectionNpv}{92.3}

\providecommand{\vitsTestfdaTriagedetectionN}{785}
\providecommand{\vitsTestfdaTriagedetectionNPos}{394}
\providecommand{\vitsTestfdaTriagedetectionNNeg}{391}
\providecommand{\vitsTestfdaTriagedetectionAuc}{0.939}
\providecommand{\vitsTestfdaTriagedetectionAucLb}{0.916}
\providecommand{\vitsTestfdaTriagedetectionAucUb}{0.960}
\providecommand{\vitsTestfdaTriagedetectionSens}{91.9}
\providecommand{\vitsTestfdaTriagedetectionSensLb}{89.1}
\providecommand{\vitsTestfdaTriagedetectionSensUb}{94.4}
\providecommand{\vitsTestfdaTriagedetectionSpec}{73.7}
\providecommand{\vitsTestfdaTriagedetectionSpecLb}{69.3}
\providecommand{\vitsTestfdaTriagedetectionSpecUb}{77.8}
\providecommand{\vitsTestfdaTriagedetectionPpv}{77.8}
\providecommand{\vitsTestfdaTriagedetectionNpv}{90.0}

\providecommand{\vitlDevEnsembleN}{2095}
\providecommand{\vitlDevEnsembleNPos}{626}
\providecommand{\vitlDevEnsembleNNeg}{1469}
\providecommand{\vitlDevEnsembleAuc}{0.922}
\providecommand{\vitlDevEnsembleAucLb}{0.903}
\providecommand{\vitlDevEnsembleAucUb}{0.938}
\providecommand{\vitlDevEnsembleSens}{84.0}
\providecommand{\vitlDevEnsembleSensLb}{80.8}
\providecommand{\vitlDevEnsembleSensUb}{86.7}
\providecommand{\vitlDevEnsembleSpec}{84.0}
\providecommand{\vitlDevEnsembleSpecLb}{82.1}
\providecommand{\vitlDevEnsembleSpecUb}{85.7}
\providecommand{\vitlDevEnsemblePpv}{69.1}
\providecommand{\vitlDevEnsembleNpv}{92.5}

\providecommand{\vitbDevEnsembleN}{2095}
\providecommand{\vitbDevEnsembleNPos}{626}
\providecommand{\vitbDevEnsembleNNeg}{1469}
\providecommand{\vitbDevEnsembleAuc}{0.917}
\providecommand{\vitbDevEnsembleAucLb}{0.898}
\providecommand{\vitbDevEnsembleAucUb}{0.934}
\providecommand{\vitbDevEnsembleSens}{83.4}
\providecommand{\vitbDevEnsembleSensLb}{80.5}
\providecommand{\vitbDevEnsembleSensUb}{86.3}
\providecommand{\vitbDevEnsembleSpec}{83.5}
\providecommand{\vitbDevEnsembleSpecLb}{81.6}
\providecommand{\vitbDevEnsembleSpecUb}{85.4}
\providecommand{\vitbDevEnsemblePpv}{68.3}
\providecommand{\vitbDevEnsembleNpv}{92.2}

\providecommand{\vitsDevEnsembleN}{2095}
\providecommand{\vitsDevEnsembleNPos}{626}
\providecommand{\vitsDevEnsembleNNeg}{1469}
\providecommand{\vitsDevEnsembleAuc}{0.913}
\providecommand{\vitsDevEnsembleAucLb}{0.895}
\providecommand{\vitsDevEnsembleAucUb}{0.931}
\providecommand{\vitsDevEnsembleSens}{82.3}
\providecommand{\vitsDevEnsembleSensLb}{79.2}
\providecommand{\vitsDevEnsembleSensUb}{85.3}
\providecommand{\vitsDevEnsembleSpec}{83.0}
\providecommand{\vitsDevEnsembleSpecLb}{81.0}
\providecommand{\vitsDevEnsembleSpecUb}{84.9}
\providecommand{\vitsDevEnsemblePpv}{67.4}
\providecommand{\vitsDevEnsembleNpv}{91.7}

\providecommand{\vitlTestfdaEnsembleN}{785}
\providecommand{\vitlTestfdaEnsembleNPos}{394}
\providecommand{\vitlTestfdaEnsembleNNeg}{391}
\providecommand{\vitlTestfdaEnsembleAuc}{0.952}
\providecommand{\vitlTestfdaEnsembleAucLb}{0.931}
\providecommand{\vitlTestfdaEnsembleAucUb}{0.969}
\providecommand{\vitlTestfdaEnsembleSens}{92.1}
\providecommand{\vitlTestfdaEnsembleSensLb}{89.3}
\providecommand{\vitlTestfdaEnsembleSensUb}{94.7}
\providecommand{\vitlTestfdaEnsembleSpec}{81.1}
\providecommand{\vitlTestfdaEnsembleSpecLb}{77.2}
\providecommand{\vitlTestfdaEnsembleSpecUb}{85.2}
\providecommand{\vitlTestfdaEnsemblePpv}{83.1}
\providecommand{\vitlTestfdaEnsembleNpv}{91.1}

\providecommand{\vitbTestfdaEnsembleN}{785}
\providecommand{\vitbTestfdaEnsembleNPos}{394}
\providecommand{\vitbTestfdaEnsembleNNeg}{391}
\providecommand{\vitbTestfdaEnsembleAuc}{0.946}
\providecommand{\vitbTestfdaEnsembleAucLb}{0.924}
\providecommand{\vitbTestfdaEnsembleAucUb}{0.965}
\providecommand{\vitbTestfdaEnsembleSens}{92.1}
\providecommand{\vitbTestfdaEnsembleSensLb}{89.3}
\providecommand{\vitbTestfdaEnsembleSensUb}{94.7}
\providecommand{\vitbTestfdaEnsembleSpec}{78.8}
\providecommand{\vitbTestfdaEnsembleSpecLb}{74.4}
\providecommand{\vitbTestfdaEnsembleSpecUb}{82.9}
\providecommand{\vitbTestfdaEnsemblePpv}{81.4}
\providecommand{\vitbTestfdaEnsembleNpv}{90.9}

\providecommand{\vitsTestfdaEnsembleN}{785}
\providecommand{\vitsTestfdaEnsembleNPos}{394}
\providecommand{\vitsTestfdaEnsembleNNeg}{391}
\providecommand{\vitsTestfdaEnsembleAuc}{0.949}
\providecommand{\vitsTestfdaEnsembleAucLb}{0.928}
\providecommand{\vitsTestfdaEnsembleAucUb}{0.968}
\providecommand{\vitsTestfdaEnsembleSens}{91.4}
\providecommand{\vitsTestfdaEnsembleSensLb}{88.6}
\providecommand{\vitsTestfdaEnsembleSensUb}{94.2}
\providecommand{\vitsTestfdaEnsembleSpec}{81.8}
\providecommand{\vitsTestfdaEnsembleSpecLb}{77.7}
\providecommand{\vitsTestfdaEnsembleSpecUb}{85.9}
\providecommand{\vitsTestfdaEnsemblePpv}{83.5}
\providecommand{\vitsTestfdaEnsembleNpv}{90.4}

\providecommand{\vitlVindrTriageN}{909}
\providecommand{\vitlVindrTriageNPos}{96}
\providecommand{\vitlVindrTriageNNeg}{813}
\providecommand{\vitlVindrTriageAuc}{0.857}
\providecommand{\vitlVindrTriageAucLb}{0.797}
\providecommand{\vitlVindrTriageAucUb}{0.910}
\providecommand{\vitlVindrTriageSens}{66.7}
\providecommand{\vitlVindrTriageSensLb}{57.3}
\providecommand{\vitlVindrTriageSensUb}{76.0}
\providecommand{\vitlVindrTriageSpec}{86.5}
\providecommand{\vitlVindrTriageSpecLb}{84.1}
\providecommand{\vitlVindrTriageSpecUb}{88.7}
\providecommand{\vitlVindrTriagePpv}{36.8}
\providecommand{\vitlVindrTriageNpv}{95.6}

\providecommand{\vitbVindrTriageN}{909}
\providecommand{\vitbVindrTriageNPos}{96}
\providecommand{\vitbVindrTriageNNeg}{813}
\providecommand{\vitbVindrTriageAuc}{0.829}
\providecommand{\vitbVindrTriageAucLb}{0.764}
\providecommand{\vitbVindrTriageAucUb}{0.889}
\providecommand{\vitbVindrTriageSens}{68.8}
\providecommand{\vitbVindrTriageSensLb}{59.4}
\providecommand{\vitbVindrTriageSensUb}{78.1}
\providecommand{\vitbVindrTriageSpec}{81.3}
\providecommand{\vitbVindrTriageSpecLb}{78.5}
\providecommand{\vitbVindrTriageSpecUb}{84.0}
\providecommand{\vitbVindrTriagePpv}{30.3}
\providecommand{\vitbVindrTriageNpv}{95.7}

\providecommand{\vitsVindrTriageN}{909}
\providecommand{\vitsVindrTriageNPos}{96}
\providecommand{\vitsVindrTriageNNeg}{813}
\providecommand{\vitsVindrTriageAuc}{0.836}
\providecommand{\vitsVindrTriageAucLb}{0.768}
\providecommand{\vitsVindrTriageAucUb}{0.896}
\providecommand{\vitsVindrTriageSens}{68.8}
\providecommand{\vitsVindrTriageSensLb}{59.4}
\providecommand{\vitsVindrTriageSensUb}{77.1}
\providecommand{\vitsVindrTriageSpec}{85.9}
\providecommand{\vitsVindrTriageSpecLb}{83.3}
\providecommand{\vitsVindrTriageSpecUb}{88.2}
\providecommand{\vitsVindrTriagePpv}{36.5}
\providecommand{\vitsVindrTriageNpv}{95.9}

\providecommand{\CnnVindrTriageN}{909}
\providecommand{\CnnVindrTriageNPos}{96}
\providecommand{\CnnVindrTriageNNeg}{813}
\providecommand{\CnnVindrTriageAuc}{0.814}
\providecommand{\CnnVindrTriageAucLb}{0.746}
\providecommand{\CnnVindrTriageAucUb}{0.879}
\providecommand{\CnnVindrTriageSens}{62.5}
\providecommand{\CnnVindrTriageSensLb}{53.1}
\providecommand{\CnnVindrTriageSensUb}{71.9}
\providecommand{\CnnVindrTriageSpec}{84.6}
\providecommand{\CnnVindrTriageSpecLb}{82.2}
\providecommand{\CnnVindrTriageSpecUb}{87.0}
\providecommand{\CnnVindrTriagePpv}{32.4}
\providecommand{\CnnVindrTriageNpv}{95.0}

\providecommand{\vitlVindrDetectionN}{909}
\providecommand{\vitlVindrDetectionNPos}{96}
\providecommand{\vitlVindrDetectionNNeg}{813}
\providecommand{\vitlVindrDetectionAuc}{0.877}
\providecommand{\vitlVindrDetectionAucLb}{0.820}
\providecommand{\vitlVindrDetectionAucUb}{0.925}
\providecommand{\vitlVindrDetectionSens}{68.8}
\providecommand{\vitlVindrDetectionSensLb}{59.4}
\providecommand{\vitlVindrDetectionSensUb}{77.1}
\providecommand{\vitlVindrDetectionSpec}{91.3}
\providecommand{\vitlVindrDetectionSpecLb}{89.3}
\providecommand{\vitlVindrDetectionSpecUb}{93.2}
\providecommand{\vitlVindrDetectionPpv}{48.2}
\providecommand{\vitlVindrDetectionNpv}{96.1}

\providecommand{\vitbVindrDetectionN}{909}
\providecommand{\vitbVindrDetectionNPos}{96}
\providecommand{\vitbVindrDetectionNNeg}{813}
\providecommand{\vitbVindrDetectionAuc}{0.859}
\providecommand{\vitbVindrDetectionAucLb}{0.797}
\providecommand{\vitbVindrDetectionAucUb}{0.913}
\providecommand{\vitbVindrDetectionSens}{65.6}
\providecommand{\vitbVindrDetectionSensLb}{56.2}
\providecommand{\vitbVindrDetectionSensUb}{75.0}
\providecommand{\vitbVindrDetectionSpec}{90.2}
\providecommand{\vitbVindrDetectionSpecLb}{88.1}
\providecommand{\vitbVindrDetectionSpecUb}{92.0}
\providecommand{\vitbVindrDetectionPpv}{44.1}
\providecommand{\vitbVindrDetectionNpv}{95.7}

\providecommand{\vitsVindrDetectionN}{909}
\providecommand{\vitsVindrDetectionNPos}{96}
\providecommand{\vitsVindrDetectionNNeg}{813}
\providecommand{\vitsVindrDetectionAuc}{0.849}
\providecommand{\vitsVindrDetectionAucLb}{0.783}
\providecommand{\vitsVindrDetectionAucUb}{0.906}
\providecommand{\vitsVindrDetectionSens}{64.6}
\providecommand{\vitsVindrDetectionSensLb}{55.2}
\providecommand{\vitsVindrDetectionSensUb}{74.0}
\providecommand{\vitsVindrDetectionSpec}{92.9}
\providecommand{\vitsVindrDetectionSpecLb}{91.0}
\providecommand{\vitsVindrDetectionSpecUb}{94.5}
\providecommand{\vitsVindrDetectionPpv}{51.7}
\providecommand{\vitsVindrDetectionNpv}{95.7}

\providecommand{\vitlVindrTriagedetectionN}{909}
\providecommand{\vitlVindrTriagedetectionNPos}{96}
\providecommand{\vitlVindrTriagedetectionNNeg}{813}
\providecommand{\vitlVindrTriagedetectionAuc}{0.871}
\providecommand{\vitlVindrTriagedetectionAucLb}{0.813}
\providecommand{\vitlVindrTriagedetectionAucUb}{0.922}
\providecommand{\vitlVindrTriagedetectionSens}{69.8}
\providecommand{\vitlVindrTriagedetectionSensLb}{61.5}
\providecommand{\vitlVindrTriagedetectionSensUb}{78.1}
\providecommand{\vitlVindrTriagedetectionSpec}{89.7}
\providecommand{\vitlVindrTriagedetectionSpecLb}{87.5}
\providecommand{\vitlVindrTriagedetectionSpecUb}{91.8}
\providecommand{\vitlVindrTriagedetectionPpv}{44.4}
\providecommand{\vitlVindrTriagedetectionNpv}{96.2}

\providecommand{\vitbVindrTriagedetectionN}{909}
\providecommand{\vitbVindrTriagedetectionNPos}{96}
\providecommand{\vitbVindrTriagedetectionNNeg}{813}
\providecommand{\vitbVindrTriagedetectionAuc}{0.851}
\providecommand{\vitbVindrTriagedetectionAucLb}{0.788}
\providecommand{\vitbVindrTriagedetectionAucUb}{0.906}
\providecommand{\vitbVindrTriagedetectionSens}{67.7}
\providecommand{\vitbVindrTriagedetectionSensLb}{58.3}
\providecommand{\vitbVindrTriagedetectionSensUb}{77.1}
\providecommand{\vitbVindrTriagedetectionSpec}{87.3}
\providecommand{\vitbVindrTriagedetectionSpecLb}{85.0}
\providecommand{\vitbVindrTriagedetectionSpecUb}{89.5}
\providecommand{\vitbVindrTriagedetectionPpv}{38.7}
\providecommand{\vitbVindrTriagedetectionNpv}{95.8}

\providecommand{\vitsVindrTriagedetectionN}{909}
\providecommand{\vitsVindrTriagedetectionNPos}{96}
\providecommand{\vitsVindrTriagedetectionNNeg}{813}
\providecommand{\vitsVindrTriagedetectionAuc}{0.844}
\providecommand{\vitsVindrTriagedetectionAucLb}{0.778}
\providecommand{\vitsVindrTriagedetectionAucUb}{0.904}
\providecommand{\vitsVindrTriagedetectionSens}{64.6}
\providecommand{\vitsVindrTriagedetectionSensLb}{55.2}
\providecommand{\vitsVindrTriagedetectionSensUb}{75.0}
\providecommand{\vitsVindrTriagedetectionSpec}{91.6}
\providecommand{\vitsVindrTriagedetectionSpecLb}{89.7}
\providecommand{\vitsVindrTriagedetectionSpecUb}{93.4}
\providecommand{\vitsVindrTriagedetectionPpv}{47.7}
\providecommand{\vitsVindrTriagedetectionNpv}{95.6}

\providecommand{\vitlVindrEnsembleN}{909}
\providecommand{\vitlVindrEnsembleNPos}{96}
\providecommand{\vitlVindrEnsembleNNeg}{813}
\providecommand{\vitlVindrEnsembleAuc}{0.871}
\providecommand{\vitlVindrEnsembleAucLb}{0.818}
\providecommand{\vitlVindrEnsembleAucUb}{0.918}
\providecommand{\vitlVindrEnsembleSens}{65.6}
\providecommand{\vitlVindrEnsembleSensLb}{56.2}
\providecommand{\vitlVindrEnsembleSensUb}{75.0}
\providecommand{\vitlVindrEnsembleSpec}{89.4}
\providecommand{\vitlVindrEnsembleSpecLb}{87.5}
\providecommand{\vitlVindrEnsembleSpecUb}{91.5}
\providecommand{\vitlVindrEnsemblePpv}{42.3}
\providecommand{\vitlVindrEnsembleNpv}{95.7}

\providecommand{\vitbVindrEnsembleN}{909}
\providecommand{\vitbVindrEnsembleNPos}{96}
\providecommand{\vitbVindrEnsembleNNeg}{813}
\providecommand{\vitbVindrEnsembleAuc}{0.865}
\providecommand{\vitbVindrEnsembleAucLb}{0.807}
\providecommand{\vitbVindrEnsembleAucUb}{0.915}
\providecommand{\vitbVindrEnsembleSens}{64.6}
\providecommand{\vitbVindrEnsembleSensLb}{55.2}
\providecommand{\vitbVindrEnsembleSensUb}{74.0}
\providecommand{\vitbVindrEnsembleSpec}{89.4}
\providecommand{\vitbVindrEnsembleSpecLb}{87.3}
\providecommand{\vitbVindrEnsembleSpecUb}{91.5}
\providecommand{\vitbVindrEnsemblePpv}{41.9}
\providecommand{\vitbVindrEnsembleNpv}{95.5}

\providecommand{\vitsVindrEnsembleN}{909}
\providecommand{\vitsVindrEnsembleNPos}{96}
\providecommand{\vitsVindrEnsembleNNeg}{813}
\providecommand{\vitsVindrEnsembleAuc}{0.849}
\providecommand{\vitsVindrEnsembleAucLb}{0.784}
\providecommand{\vitsVindrEnsembleAucUb}{0.906}
\providecommand{\vitsVindrEnsembleSens}{62.5}
\providecommand{\vitsVindrEnsembleSensLb}{53.1}
\providecommand{\vitsVindrEnsembleSensUb}{71.9}
\providecommand{\vitsVindrEnsembleSpec}{90.4}
\providecommand{\vitsVindrEnsembleSpecLb}{88.3}
\providecommand{\vitsVindrEnsembleSpecUb}{92.3}
\providecommand{\vitsVindrEnsemblePpv}{43.5}
\providecommand{\vitsVindrEnsembleNpv}{95.3}

%% file: results/combined/delong/variables.tex
\providecommand{\delongVitbVsCnnPval}{0.90}
\providecommand{\delongVitbVsCnnPvalRaw}{0.897586}
\providecommand{\delongVitbVsCnnAucOne}{0.9235}
\providecommand{\delongVitbVsCnnAucTwo}{0.9247}
\providecommand{\delongVitbVsCnnAucDiff}{-0.0012}
\providecommand{\delongVitbVsCnnAucDiffLb}{-0.0193}
\providecommand{\delongVitbVsCnnAucDiffUb}{0.0169}
\providecommand{\delongVitbVsCnnSeDiff}{0.0092}
\providecommand{\delongVitbVsCnnZstat}{-0.1287}
\providecommand{\delongVitbVsCnnNsamples}{785}
\providecommand{\delongVitbVsCnnNpositive}{394}
\providecommand{\delongVitbVsCnnNnegative}{391}
\providecommand{\delongVitbVsCnnSig}{}
\providecommand{\delongVitbVsCnnAucDiffCi}{-0.0012 [-0.0193--0.0169]}
\providecommand{\delongVitbVsCnnModelOne}{ViT-B}
\providecommand{\delongVitbVsCnnModelTwo}{CNN}

\providecommand{\delongVitlVsCnnPval}{0.02}
\providecommand{\delongVitlVsCnnPvalRaw}{0.016487}
\providecommand{\delongVitlVsCnnAucOne}{0.9450}
\providecommand{\delongVitlVsCnnAucTwo}{0.9247}
\providecommand{\delongVitlVsCnnAucDiff}{0.0203}
\providecommand{\delongVitlVsCnnAucDiffLb}{0.0037}
\providecommand{\delongVitlVsCnnAucDiffUb}{0.0370}
\providecommand{\delongVitlVsCnnSeDiff}{0.0085}
\providecommand{\delongVitlVsCnnZstat}{2.3979}
\providecommand{\delongVitlVsCnnNsamples}{785}
\providecommand{\delongVitlVsCnnNpositive}{394}
\providecommand{\delongVitlVsCnnNnegative}{391}
\providecommand{\delongVitlVsCnnSig}{*}
\providecommand{\delongVitlVsCnnAucDiffCi}{0.0203 [0.0037--0.0370]}
\providecommand{\delongVitlVsCnnModelOne}{ViT-L}
\providecommand{\delongVitlVsCnnModelTwo}{CNN}

\providecommand{\delongVitsVsCnnPval}{0.25}
\providecommand{\delongVitsVsCnnPvalRaw}{0.252040}
\providecommand{\delongVitsVsCnnAucOne}{0.9350}
\providecommand{\delongVitsVsCnnAucTwo}{0.9247}
\providecommand{\delongVitsVsCnnAucDiff}{0.0103}
\providecommand{\delongVitsVsCnnAucDiffLb}{-0.0073}
\providecommand{\delongVitsVsCnnAucDiffUb}{0.0278}
\providecommand{\delongVitsVsCnnSeDiff}{0.0090}
\providecommand{\delongVitsVsCnnZstat}{1.1454}
\providecommand{\delongVitsVsCnnNsamples}{785}
\providecommand{\delongVitsVsCnnNpositive}{394}
\providecommand{\delongVitsVsCnnNnegative}{391}
\providecommand{\delongVitsVsCnnSig}{}
\providecommand{\delongVitsVsCnnAucDiffCi}{0.0103 [-0.0073--0.0278]}
\providecommand{\delongVitsVsCnnModelOne}{ViT-S}
\providecommand{\delongVitsVsCnnModelTwo}{CNN}

%% file: results/vit-l/test/density/multiclass/latex/variables.tex
\newcommand{\vitlDensityAccuracyMean}{0.626}
\newcommand{\vitlDensityAccuracyCILower}{0.607}
\newcommand{\vitlDensityAccuracyCIUpper}{0.645}
\newcommand{\vitlDensityAdjacentAccuracyMean}{0.988}
\newcommand{\vitlDensityAdjacentAccuracyCILower}{0.984}
\newcommand{\vitlDensityAdjacentAccuracyCIUpper}{0.992}

%% file: results/vit-l/test/density/binary/latex/variables.tex
\providecommand{\vitlDensityBinarySamples}{2262}
\providecommand{\vitlDensityBinaryPositive}{1110}
\providecommand{\vitlDensityBinaryNegative}{1152}
\providecommand{\vitlDensityBinaryAuc}{0.9529}
\providecommand{\vitlDensityBinaryAucLb}{0.9401}
\providecommand{\vitlDensityBinaryAucUb}{0.9641}
\providecommand{\vitlDensityBinaryTp}{1078}
\providecommand{\vitlDensityBinaryFp}{317}
\providecommand{\vitlDensityBinaryTn}{835}
\providecommand{\vitlDensityBinaryFn}{32}
\providecommand{\vitlDensityBinaryNpv}{0.9631}
\providecommand{\vitlDensityBinaryPpv}{0.7728}
\providecommand{\vitlDensityBinaryAccuracy}{0.8457}
\providecommand{\vitlDensityBinarySensitivity}{0.9712}
\providecommand{\vitlDensityBinarySpecificity}{0.7248}
\providecommand{\vitlDensityBinarySensitivityLb}{0.9604}
\providecommand{\vitlDensityBinarySensitivityUb}{0.9811}
\providecommand{\vitlDensityBinarySpecificityLb}{0.6988}
\providecommand{\vitlDensityBinarySpecificityUb}{0.7526}

%% file: results/vit-b/test/density/multiclass/latex/variables.tex
\newcommand{\vitbDensityAccuracyMean}{0.621}
\newcommand{\vitbDensityAccuracyCILower}{0.602}
\newcommand{\vitbDensityAccuracyCIUpper}{0.639}
\newcommand{\vitbDensityAdjacentAccuracyMean}{0.987}
\newcommand{\vitbDensityAdjacentAccuracyCILower}{0.982}
\newcommand{\vitbDensityAdjacentAccuracyCIUpper}{0.991}

%% file: results/vit-b/test/density/binary/latex/variables.tex
\providecommand{\vitbDensityBinarySamples}{2262}
\providecommand{\vitbDensityBinaryPositive}{1110}
\providecommand{\vitbDensityBinaryNegative}{1152}
\providecommand{\vitbDensityBinaryAuc}{0.9524}
\providecommand{\vitbDensityBinaryAucLb}{0.9394}
\providecommand{\vitbDensityBinaryAucUb}{0.9635}
\providecommand{\vitbDensityBinaryTp}{1080}
\providecommand{\vitbDensityBinaryFp}{333}
\providecommand{\vitbDensityBinaryTn}{819}
\providecommand{\vitbDensityBinaryFn}{30}
\providecommand{\vitbDensityBinaryNpv}{0.9647}
\providecommand{\vitbDensityBinaryPpv}{0.7643}
\providecommand{\vitbDensityBinaryAccuracy}{0.8395}
\providecommand{\vitbDensityBinarySensitivity}{0.9730}
\providecommand{\vitbDensityBinarySpecificity}{0.7109}
\providecommand{\vitbDensityBinarySensitivityLb}{0.9631}
\providecommand{\vitbDensityBinarySensitivityUb}{0.9829}
\providecommand{\vitbDensityBinarySpecificityLb}{0.6832}
\providecommand{\vitbDensityBinarySpecificityUb}{0.7370}

%% file: results/vit-s/test/density/multiclass/latex/variables.tex
\newcommand{\vitsDensityAccuracyMean}{0.605}
\newcommand{\vitsDensityAccuracyCILower}{0.586}
\newcommand{\vitsDensityAccuracyCIUpper}{0.624}
\newcommand{\vitsDensityAdjacentAccuracyMean}{0.987}
\newcommand{\vitsDensityAdjacentAccuracyCILower}{0.983}
\newcommand{\vitsDensityAdjacentAccuracyCIUpper}{0.991}

%% file: results/vit-s/test/density/binary/latex/variables.tex
\providecommand{\vitsDensityBinarySamples}{2262}
\providecommand{\vitsDensityBinaryPositive}{1110}
\providecommand{\vitsDensityBinaryNegative}{1152}
\providecommand{\vitsDensityBinaryAuc}{0.9526}
\providecommand{\vitsDensityBinaryAucLb}{0.9401}
\providecommand{\vitsDensityBinaryAucUb}{0.9641}
\providecommand{\vitsDensityBinaryTp}{1080}
\providecommand{\vitsDensityBinaryFp}{333}
\providecommand{\vitsDensityBinaryTn}{819}
\providecommand{\vitsDensityBinaryFn}{30}
\providecommand{\vitsDensityBinaryNpv}{0.9647}
\providecommand{\vitsDensityBinaryPpv}{0.7643}
\providecommand{\vitsDensityBinaryAccuracy}{0.8395}
\providecommand{\vitsDensityBinarySensitivity}{0.9730}
\providecommand{\vitsDensityBinarySpecificity}{0.7109}
\providecommand{\vitsDensityBinarySensitivityLb}{0.9631}
\providecommand{\vitsDensityBinarySensitivityUb}{0.9820}
\providecommand{\vitsDensityBinarySpecificityLb}{0.6858}
\providecommand{\vitsDensityBinarySpecificityUb}{0.7370}

%% file: results/vit-l/vindr/density/multiclass/latex/variables.tex
\newcommand{\vitlVindrDensityAccuracyMean}{0.276}
\newcommand{\vitlVindrDensityAccuracyCILower}{0.248}
\newcommand{\vitlVindrDensityAccuracyCIUpper}{0.306}
\newcommand{\vitlVindrDensityAdjacentAccuracyMean}{0.989}
\newcommand{\vitlVindrDensityAdjacentAccuracyCILower}{0.981}
\newcommand{\vitlVindrDensityAdjacentAccuracyCIUpper}{0.996}

%% file: results/vit-l/vindr/density/binary/latex/variables.tex
\providecommand{\vitlVindrDensityBinarySamples}{909}
\providecommand{\vitlVindrDensityBinaryPositive}{814}
\providecommand{\vitlVindrDensityBinaryNegative}{95}
\providecommand{\vitlVindrDensityBinaryAuc}{0.9545}
\providecommand{\vitlVindrDensityBinaryAucLb}{0.9268}
\providecommand{\vitlVindrDensityBinaryAucUb}{0.9760}
\providecommand{\vitlVindrDensityBinaryTp}{803}
\providecommand{\vitlVindrDensityBinaryFp}{51}
\providecommand{\vitlVindrDensityBinaryTn}{44}
\providecommand{\vitlVindrDensityBinaryFn}{11}
\providecommand{\vitlVindrDensityBinaryNpv}{0.8000}
\providecommand{\vitlVindrDensityBinaryPpv}{0.9403}
\providecommand{\vitlVindrDensityBinaryAccuracy}{0.9318}
\providecommand{\vitlVindrDensityBinarySensitivity}{0.9865}
\providecommand{\vitlVindrDensityBinarySpecificity}{0.4632}
\providecommand{\vitlVindrDensityBinarySensitivityLb}{0.9779}
\providecommand{\vitlVindrDensityBinarySensitivityUb}{0.9939}
\providecommand{\vitlVindrDensityBinarySpecificityLb}{0.3579}
\providecommand{\vitlVindrDensityBinarySpecificityUb}{0.5579}

%% file: results/vit-b/vindr/density/multiclass/latex/variables.tex
\newcommand{\vitbVindrDensityAccuracyMean}{0.272}
\newcommand{\vitbVindrDensityAccuracyCILower}{0.244}
\newcommand{\vitbVindrDensityAccuracyCIUpper}{0.300}
\newcommand{\vitbVindrDensityAdjacentAccuracyMean}{0.987}
\newcommand{\vitbVindrDensityAdjacentAccuracyCILower}{0.979}
\newcommand{\vitbVindrDensityAdjacentAccuracyCIUpper}{0.994}

%% file: results/vit-b/vindr/density/binary/latex/variables.tex
\providecommand{\vitbVindrDensityBinarySamples}{909}
\providecommand{\vitbVindrDensityBinaryPositive}{814}
\providecommand{\vitbVindrDensityBinaryNegative}{95}
\providecommand{\vitbVindrDensityBinaryAuc}{0.9566}
\providecommand{\vitbVindrDensityBinaryAucLb}{0.9307}
\providecommand{\vitbVindrDensityBinaryAucUb}{0.9773}
\providecommand{\vitbVindrDensityBinaryTp}{806}
\providecommand{\vitbVindrDensityBinaryFp}{47}
\providecommand{\vitbVindrDensityBinaryTn}{48}
\providecommand{\vitbVindrDensityBinaryFn}{8}
\providecommand{\vitbVindrDensityBinaryNpv}{0.8571}
\providecommand{\vitbVindrDensityBinaryPpv}{0.9449}
\providecommand{\vitbVindrDensityBinaryAccuracy}{0.9395}
\providecommand{\vitbVindrDensityBinarySensitivity}{0.9902}
\providecommand{\vitbVindrDensityBinarySpecificity}{0.5053}
\providecommand{\vitbVindrDensityBinarySensitivityLb}{0.9828}
\providecommand{\vitbVindrDensityBinarySensitivityUb}{0.9963}
\providecommand{\vitbVindrDensityBinarySpecificityLb}{0.4105}
\providecommand{\vitbVindrDensityBinarySpecificityUb}{0.6105}

%% file: results/vit-s/vindr/density/multiclass/latex/variables.tex
\newcommand{\vitsVindrDensityAccuracyMean}{0.276}
\newcommand{\vitsVindrDensityAccuracyCILower}{0.250}
\newcommand{\vitsVindrDensityAccuracyCIUpper}{0.307}
\newcommand{\vitsVindrDensityAdjacentAccuracyMean}{0.991}
\newcommand{\vitsVindrDensityAdjacentAccuracyCILower}{0.985}
\newcommand{\vitsVindrDensityAdjacentAccuracyCIUpper}{0.997}

%% file: results/vit-s/vindr/density/binary/latex/variables.tex
\providecommand{\vitsVindrDensityBinarySamples}{909}
\providecommand{\vitsVindrDensityBinaryPositive}{814}
\providecommand{\vitsVindrDensityBinaryNegative}{95}
\providecommand{\vitsVindrDensityBinaryAuc}{0.9599}
\providecommand{\vitsVindrDensityBinaryAucLb}{0.9360}
\providecommand{\vitsVindrDensityBinaryAucUb}{0.9795}
\providecommand{\vitsVindrDensityBinaryTp}{800}
\providecommand{\vitsVindrDensityBinaryFp}{40}
\providecommand{\vitsVindrDensityBinaryTn}{55}
\providecommand{\vitsVindrDensityBinaryFn}{14}
\providecommand{\vitsVindrDensityBinaryNpv}{0.7971}
\providecommand{\vitsVindrDensityBinaryPpv}{0.9524}
\providecommand{\vitsVindrDensityBinaryAccuracy}{0.9406}
\providecommand{\vitsVindrDensityBinarySensitivity}{0.9828}
\providecommand{\vitsVindrDensityBinarySpecificity}{0.5789}
\providecommand{\vitsVindrDensityBinarySensitivityLb}{0.9730}
\providecommand{\vitsVindrDensityBinarySensitivityUb}{0.9902}
\providecommand{\vitsVindrDensityBinarySpecificityLb}{0.4842}
\providecommand{\vitsVindrDensityBinarySpecificityUb}{0.6842}